\documentclass[sigconf]{acmart}
\usepackage{graphicx}
\usepackage{subfigure}
\usepackage{svg}
\usepackage[countmax]{subfloat}
\usepackage{subfloat}
\usepackage{subcaption}
\usepackage{booktabs} 
\usepackage{tabularx}
\usepackage{threeparttable}
\usepackage{multirow}
\usepackage{amsmath}
\usepackage{subcaption}
\usepackage{makecell}
\usepackage{enumerate}
\usepackage{enumitem}
\usepackage{hyperref}
\usepackage{algorithm}
\usepackage{algpseudocode}
\usepackage{dblfloatfix}



\usepackage{tabularx}
\AtBeginDocument{%
  }

\setcopyright{acmlicensed}
\copyrightyear{2027}
\acmYear{2027}
\acmDOI{XXXXXXX.XXXXXXX}
\acmConference[Conference acronym 'XX]{Make sure to enter the correct
  conference title from your rights confirmation email}{June 03--05,
  2027}{Woodstock, NY}

\begin{document}

\newcommand{\M}{AdaMamba}
\title[\M]{\M: An Adaptive State Space Model for Accommodating Time-Frequency Heterogeneity in Long-Term Time Series Forecasting}



\author{Xudong Jiang}
\affiliation{%
  \institution{Tongji University}
  \city{Shanghai}
  \country{China}
}
\email{xdjiang@tongji.edu.cn}

\author{Mingshan LOO}
\affiliation{%
  \institution{Tongji University}
  \city{Shanghai}
  \country{China}
}
\email{2493055@tongji.edu.cn}

\author{Wengen Li}
\authornotemark[1]
\affiliation{%
  \institution{Tongji University}
  \city{Shanghai}
  \country{China}
}
\email{lwengen@tongji.edu.cn}

\author{Mingrui Zhang}
\affiliation{%
  \institution{Tongji University}
  \city{Shanghai}
  \country{China}
}
\email{2534017@tongji.edu.cn}

\author{Jihong Guan}
\authornotemark[1]
\affiliation{%
  \institution{Tongji University}
  \city{Shanghai}
  \country{China}
}
\email{jhguan@tongji.edu.cn}

\author{Shuigeng Zhou}
\affiliation{%
  \institution{Fudan University}
  \city{Shanghai}
  \country{China}
}
\email{sgzhou@fudan.edu.cn}

\renewcommand{\shortauthors}{Xudong Jiang et al.}

\begin{abstract}
Time-series forecasting is foundational to a wide spectrum of web-scale services, spanning personalized recommendations, energy scheduling, and infrastructure capacity planning.
Accurate long-term time series forecasting (LTSF) requires models to capture long-range temporal dependencies while adapting to periodic patterns that vary across variables. 
Although frequency-aware methods offer a global view of temporal dynamics, they typically rely on inflexible heuristic designs, such as predefined period lengths, manually selected or randomly sampled top-$k$ frequency components, and separate spectral transformations. Consequently, they fail to account for time-frequency heterogeneity, where variables that appear synchronized in the time domain can differ substantially in the frequency domain. 
Meanwhile, Mamba-based forecasters, despite their efficiency in long-range modeling, introduce frequency information only as auxiliary signals, leaving spectral evidence disconnected from state transitions. 
In this work, we propose \M, a novel approach that explicitly internalizes input-dependent frequency information into the modeling of temporal dependencies.
\M~comprises an interaction learning module and an adaptive frequency-gated state space module: the former captures cross-variable dependencies and multi-scale temporal context, while the latter generates context-dependent frequency bases and unifies temporal forgetting into a single time-frequency gate that directly modulates state evolution. Extensive experiments on seven public LTSF benchmarks and four domain-specific datasets demonstrate that \M~consistently outperforms state-of-the-art baselines while maintaining competitive computational efficiency.
The code is available at~\url{https://github.com/XDjiang25/AdaMamba}.

\end{abstract}




\maketitle

\section{Introduction}
The modern Web is a continuously evolving ecosystem in which services generate large streams of behavioral and operational data, ranging from user interactions and recommendation feedback to energy telemetry and system logs~\cite{wen2022robust,lim2021time}. 
Accurate forecasting of such streams underpins core application scenarios such as personalized content recommendation, online economic modeling, and infrastructure resource allocation~\cite{barigozzi2019nets,zhang2015daily,liu2022scinet}, and enables platforms to respond proactively by anticipating traffic surges, predicting engagement trends, and detecting anomalies before service degradation. As many of these scenarios require planning over extended horizons, long-term time-series forecasting (LTSF) 
thus becomes a foundational capability for modern Web applications.

A fundamental challenge of LTSF is to capture long-range temporal dependencies while remaining sensitive to dynamic periodic patterns. 
Frequency-domain analysis is particularly appealing in this context because it reveals global temporal structures that are difficult to identify from raw observations alone~\cite{li2025faith,wang2025filterts}. 


\begin{figure}[ht]
 \centering {\includegraphics[width=1\linewidth]{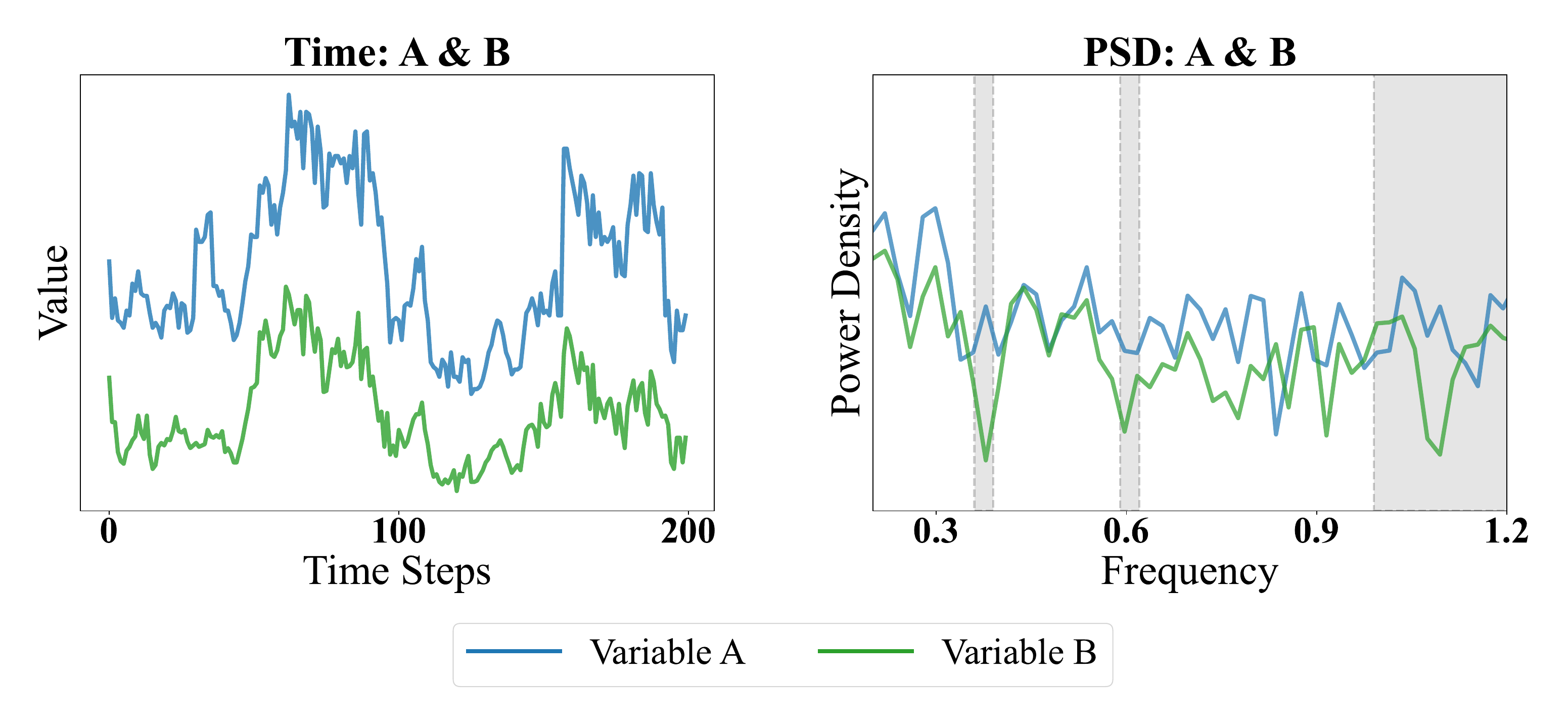}}%
 \hfil
 \centering
 \caption{A motivating example of the ETTm1 dataset in both time and frequency domains. 
 While ``variable A'' \& ``variable B'',
 exhibit noticeable similarities in the time domain, their frequency characteristics differ markedly, particularly in the gray-highlighted regions. (Left) Time domain: time-series variations over 200 time steps. (Right) Frequency domain: frequency discrepancies visualized via PSD.
 }
 \label{fig_Introduction}
\end{figure}

\begin{figure*}[!b]
  \centering
  \includegraphics[width=\linewidth]{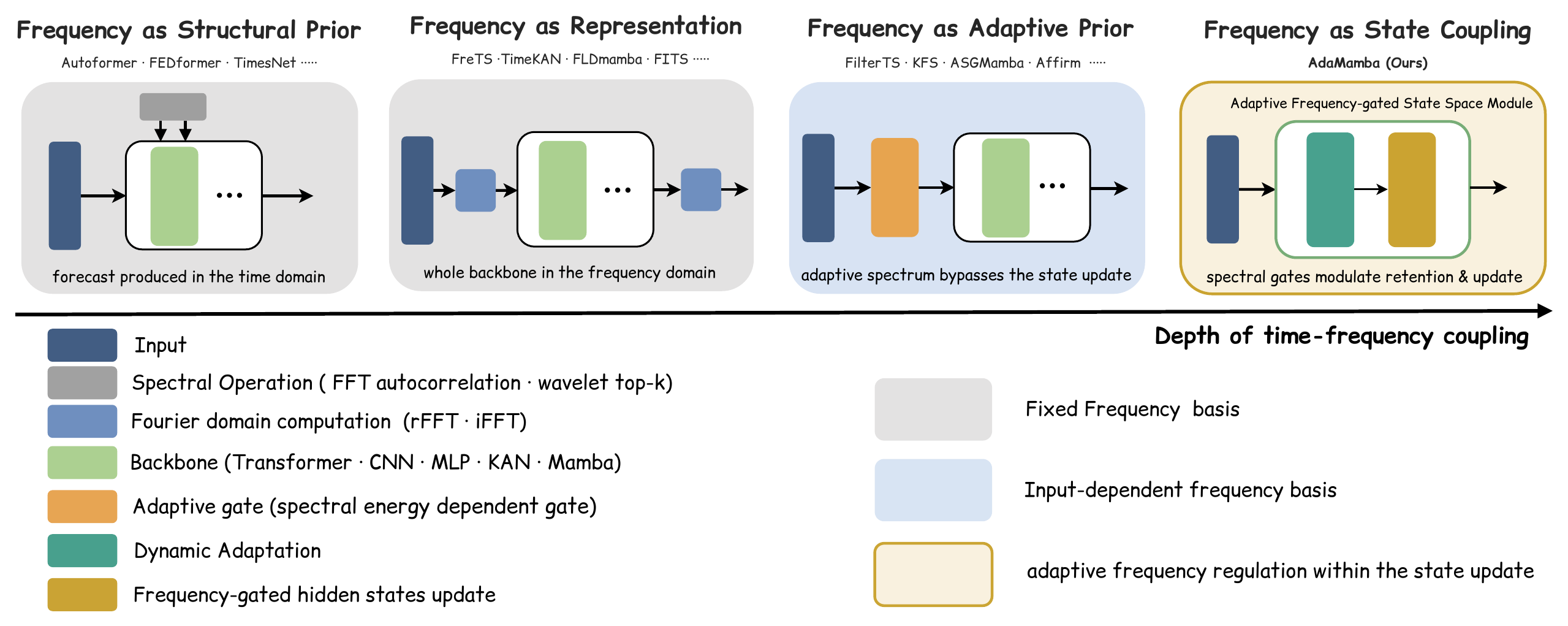}
  \caption{Comparison between \M~and closely related frequency-aware methods, organized by how deeply spectral information is coupled into the model. The arrow indicates increasing coupling depth; the fixed vs.\ input-dependent basis is annotated for each family.
  }
  \label{fig:relatedwork}
\end{figure*}

However, exploiting frequency information for LTSF is non-trivial. In real-world time series, temporal similarity and spectral structure are often misaligned: variables that appear similar in the time domain can exhibit substantially different power spectral density (PSD) patterns. As illustrated in Fig.~\ref{fig_Introduction}, several pairs of variables in the ETTm1 dataset show nearly synchronized variations over time, yet their spectra differ markedly in the gray-highlighted regions. We term this discrepancy \textit{time-frequency heterogeneity}: temporal synchrony does not guarantee frequency consistency, and the relevant periodic structure can vary across variables and inputs. 
Consequently, effective LTSF calls for frequency analysis that adapts to the input, rather than relying on fixed or heuristic spectral designs.


Unfortunately, most existing frequency-aware LTSF methods operate under an implicit assumption of time-frequency homogeneity. 
They commonly depend on inflexible heuristics, such as predefined period lengths~\cite{zhang2024frnet}, manually selected or randomly sampled top‑$k$ frequency components~\cite{zhou2024sdformer,wu2022timesnet}, and separate spectral transformations~\cite{huang2025timekan, DLinear}. 
Such designs are effective when dominant periodic patterns are stable, but offer limited flexibility when frequency characteristics shift across variables, samples, and forecasting horizons. 
Meanwhile, although Mamba-based forecasters provide efficient long-range modeling, they operate strictly in the time domain, and frequency information is commonly introduced through decomposition, external filtering, or branch-level processing~\cite{STDMamba,wu2025affirm,zhang2025fldmamba,guan2026slan}. Since these spectral operations are detached from the state transitions that determine what historical information is retained or forgotten, learned spectral cues cannot directly regulate state evolution.


Our key insight from these observations is that time-frequency heterogeneity should be modeled at the state-transition level, i.e., using input-dependent frequency content to directly regulate state retention, attenuation, and update. 
To this end, we propose \M, a dedicated state space model that integrates adaptive frequency regulation into the Mamba update. 
Specifically, an interaction learning module first captures cross-variable dependencies and multi-scale temporal context. 
An adaptive frequency-gated state space module then generates context-dependent frequency bases and extends temporal forgetting into a unified time-frequency gate, allowing learned spectral importance to directly modulate state evolution rather than serving merely as preprocessing or an auxiliary branch.

In summary, our major contributions are as follows:
\begin{itemize}[topsep=8pt, itemsep=4pt, parsep=0pt, partopsep=0pt]
    \item We identify time-frequency heterogeneity as a key obstacle in LTSF and propose \M, which integrates adaptive frequency regulation directly into the state space mechanism, enabling seamless joint learning of long-range temporal dependencies and dynamic periodic patterns. 
    \item We design an adaptive frequency-gated state space module that generates input-conditioned frequency bases
    and extends temporal forgetting into a unified time-frequency gate. Unlike designs that treat spectral and temporal modeling separately, this gate directly modulates state retention, attenuation, and update, so spectral information governs state evolution rather than serving as preprocessing or an auxiliary branch.
    \item {We conduct extensive experiments on seven public benchmarks and four domain-specific datasets against Mamba-based, frequency-domain, Transformer-based, MLP-based, and convolutional baselines. \M~delivers consistent improvements at competitive computational cost.} 
\end{itemize}

\section{Related Work}

\subsection{Frequency-aware Long-term Time Series Forecasting}
\label{sec:ltsf}

Growing evidence further suggests that time-series observations can be effectively modeled as compositions of complex signals evolving over time~\cite{elvander2020defining}. Analyzing these signals in the frequency domain uncovers hidden periodic patterns that are elusive in the raw temporal representation, often at a lower parameter cost. 
Existing frequency-aware methods can be grouped by how the spectrum enters the model. 
Early work exploited frequency information primarily as a structural prior to guide temporal modeling: Autoformer~\cite{wu2021autoformer} applies FFT-based autocorrelation to discover periodic dependencies, FEDformer~\cite{zhou2022fedformer} sparsifies attention with Fourier and wavelet bases, and TimesNet~\cite{wu2022timesnet} reshapes 1D series into 2D tensors according to the top-$k$ dominant periods. In this family, the spectrum serves as an auxiliary operation that determines how dependencies are discovered, attended to, or organized, whereas the forecast itself is still produced in the time domain. 

Beyond guiding how temporal structure is organized, a parallel line of work places frequency directly into the representation and the prediction. FreTS~\cite{yi2023frets} casts the entire MLP into the frequency domain, FITS~\cite{xu2024fits} formulates forecasting as frequency interpolation through a complex-valued linear layer, and FreDF~\cite{wang2024fredf} leaves the architecture untouched but supervises predictions with a spectral loss. Related variants further diversify how frequency decomposition is performed, with TimeKAN~\cite{huang2025timekan} operating through Kolmogorov--Arnold Networks (KAN), FAITH~\cite{li2025faith} through dual-horizon attention, and MixLinear~\cite{ma2026mixlinear} through lightweight linear mixing, while FRWKV~\cite{yang2025frwkv} and FreqLite~\cite{baig2026freqlite} migrate frequency-domain computation into linear attention and lightweight backbones. 
A more recent line of work relaxes the fixed-frequency premise by learning the spectral filter. FiLM~\cite{zhou2022film} projects sequences onto Legendre--Fourier bases with a learnable low-pass; FilterNet~\cite{yi2024filternet} trains neural filters (PaiFilter/TexFilter) in front of an MLP; ButterMamba~\cite{zhang2026buttermamba} inserts Butterworth-style filtering before a Mamba encoder; Filter-then-Attend~\cite{dayag2026filter} and FreqLens~\cite{chen2026freqlens} learn spectral weightings jointly with the forecaster. FilterTS~\cite{wang2025filterts} and KFS~\cite{wu2026kfs} go one step further and generate the filter dynamically from each input, achieving an input-dependent frequency basis.

Yet a ceiling remains: every filter in this family, static or dynamic, operates on representations. It alters what the model observes, whereas the state transition that determines how historical information is retained, attenuated, and updated remains frequency-agnostic. The fixed-frequency assumption discussed above is thus merely relocated from the basis to the filter. Consequently, variables that are synchronized in the time domain yet divergent in the spectrum (Fig.~\ref{fig_Introduction}) are still processed through identical temporal dynamics, leaving the time-frequency heterogeneity that motivates our work unresolved.

\subsection{State Space Models}
\label{sec:ssm}
State space models (SSMs), especially Mamba~\cite{Mamba}, are a competitive alternative to Transformers for LTSF: 
their data-dependent selective mechanism maintains a compressed latent state for long-range dependencies at linear cost. 
Building upon these advantages, numerous Mamba-based forecasters have been tailored to capture complex temporal dynamics.
For instance, STDMamba~\cite{STDMamba} combines Gaussian decomposition with Mamba to capture long-range dependencies, 
S-Mamba~\cite{wang2025smamba} employs a Mamba-based encoding layer to capture cross-variable correlations, 
and TimeMachine~\cite{ahamed2024timemachine} utilizes a multi-Mamba architecture to selectively integrate multi-scale global and local contexts. 
Despite their success, the majority of existing Mamba-based models operate strictly in the time domain. 
To alleviate this constraint, recent efforts have begun to incorporate spectral information into the Mamba architecture. 
For example, FLDmamba~\cite{zhang2025fldmamba} and Affirm~\cite{wu2025affirm} introduce Fourier-based filtering into Mamba-style blocks, while concurrent work ASGMamba~\cite{li2026asgmamba} leverages patch-level spectral energy to construct adaptive gates that modulate inputs prior to Mamba encoding.

Nevertheless, in all these cases the spectral operations remain decoupled from the state transition: they alter what is fed into the Mamba block, but they do not govern how historical information is retained, attenuated, and updated inside the block. 
Such a loosely coupled integration cannot accommodate time-frequency heterogeneity. As summarized in Fig.~\ref{fig:relatedwork}, it leaves both design dimensions underexplored: the frequency basis is not fully context-dependent, and the frequency--model coupling never reaches the state transition. 
\M~closes this gap by integrating adaptive frequency regulation directly into the state space mechanism, thereby advancing the two dimensions simultaneously.

\subsection{Discussion}
Fig.~\ref{fig:relatedwork} organizes prior work along two dimensions: how the frequency basis is obtained and how deeply spectral information is coupled into the model, with methods ordered by coupling depth. 
Existing methods either use a fixed basis as a structural prior and forecast in the time domain, 
use a fixed basis as the backbone representation, or learn input-dependent bases that still bypass state updates. 
To our knowledge, \M~is the first to carry the frequency basis into the state transition, where context-dependent frequency parameters drive trigonometric modulation and time-frequency gating to regulate state updates directly.

\section{Methodology}
\subsection{Problem Definition}
Given the historical time series $\mathcal{X}^{\mathrm{Input}}=\{x_{t-T+1},\dots,x_{t-1},x_{t}\}\in\mathbb{R}^{T\times D}$, where $T$ is the number of time steps, $D$ is the number of variables, and $x_{t}=\{x_{t,1},x_{t,2},\ldots,x_{t,D}\}\in\mathbb{R}^{1\times D}$ denotes the observed values of all $D$ variables at time step $t$, the LTSF task aims to learn a nonlinear mapping function $\mathcal{F}$ to predict the future values $\mathcal{X}^{\mathrm{Output}}=\{x_{t+1},x_{t+2},\dots,x_{t+H}\}\in\mathbb{R}^{H\times D}$ over the next $H$ time steps, i.e.,
\begin{equation}
\label{equ_1}
\mathcal{X}^{\mathrm{Output}}=\mathcal{F}_{\theta}(\mathcal{X}^{\mathrm{Input}}),
\end{equation}
where $H$ is typically large (e.g., $H \ge 96$ in power deployment scenarios), and $\theta$ denotes all the learnable parameters in $\mathcal{F}$.

\subsection{Framework Overview}
\begin{figure*}[h!]
  \centering
  \includegraphics[width=\linewidth]{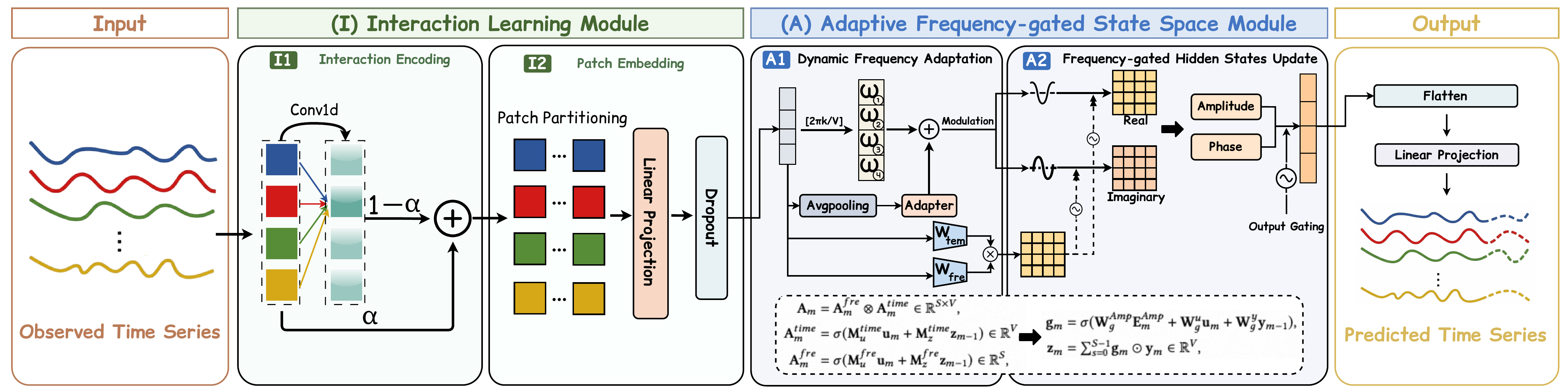}
  \caption{The architecture of \M, which comprises two key components: an interaction learning module and an adaptive frequency-gated state space module. The interaction learning module consists of an interaction encoding sub-layer and a patch embedding sub-layer, which respectively capture informative inter-variable relationships and scale-specific patch representation, while the adaptive frequency-gated state space module performs context-dependent frequency adaptation via a lightweight adapter and extends temporal forgetting into a unified time-frequency gate that directly modulates state evolution. 
  }
\label{fig:architecture}
\end{figure*}

Fig.~\ref{fig:architecture} shows the architecture of \M, which integrates adaptive frequency analysis with long-range dependency modeling. Given the input series, the interaction learning module first encodes inter-variable interactions and produces multiscale patch representations. The adaptive frequency-gated state space module then internalizes adaptive and context-aware frequency analysis into the Mamba state space update, extending temporal forgetting into a unified time-frequency gate that filters input-relevant information in a frequency-gated manner. We elaborate on the two modules in the following subsections.



\subsection{Interaction Learning Module}
\subsubsection{Interaction Encoding}
Properly capturing inter-variable correlations prior to time-dependent modeling can enhance LTSF accuracy~\cite{wang2025csformer}. To this end, given the input time series $\mathcal{X}^{\mathrm{Input}}\in\mathbb{R}^{T\times D}$ defined in Eq.~\eqref{equ_1}, we introduce interaction encoding based on a Conv1D operation to extract correlations between variables, i.e.,

\begin{equation}
\label{Interaction}
\mathcal{X}^{\mathrm{Inter}}=\text{Conv1D}(\mathcal{X}^{\mathrm{Input}}), \quad \mathcal{X}=\alpha\,\mathcal{X}^{\mathrm{Inter}}+(1-\alpha)\,\mathcal{X}^{\mathrm{Input}},
\end{equation}
where $\alpha$ is a learnable parameter that controls the relative contribution of the raw and interaction-enhanced signals.

\subsubsection{Patch Embedding}
For the output of interaction encoding, i.e., $\mathcal{X} \in {{\mathbb{R}}^{T\times D}}$, we adopt a channel-independent strategy to obtain semantically informative representations.   

First, we define a set of patch lengths $P=\left \{P_{1},...,P_{K}\right \}$.
For each $P_{i} \in P$, a patch partitioning operation with stride $s$ divides $\mathcal{X}$ into $Q$ patches $\mathcal{X}_{1},...,\mathcal{X}_{Q}$, where scale-specific padding is applied to keep $Q$ consistent across all patch lengths. 
Then, each patch $\mathcal{X}_{i}\in {\mathbb{R}}^{D \times P_{i}} $ is projected to $\mathcal{X}^{embed}_{i}\in {\mathbb{R}}^{D \times V}$ from its original dimension $P_{i}$ to a unified hidden dimension $V$ through a learnable linear projection, i.e.,
\begin{equation}
\label{equ_2}
\mathcal{X}^{embed}_{i}=\text{PatchEmbed} (\mathcal{X}_{i} ).
\end{equation}

Thus, for patch length $P_i$, we have $\mathcal{X}^{P_{i}}=\left \{ \mathcal{X}^{embed}_{1},...,\mathcal{X}^{embed}_{Q} \right \} \in {\mathbb{R}}^{D \times Q \times V}$. 
Similarly, the process is applied independently for each $P_i \in P$, and the embeddings from all patch scales are concatenated along the patch dimension to form the final multiscale representation:
\begin{equation}
\mathbf{U}=\text{Concat}(\mathcal{X}_{}^{P_{1}},...,\mathcal{X}_{}^{P_{K}} ) \in {\mathbb{R}}^{D \times M \times V},
\end{equation}
where $M=K \times Q$. 
Different patch lengths provide diverse temporal receptive fields, thereby enabling \M~to capture both fine-grained local variations and global trends. 

\subsection{Adaptive Frequency-Gated State Space Module}
\subsubsection{Dynamic Frequency Adaptation}

In contrast to conventional Fourier analysis with a fixed frequency grid, we introduce dynamically adaptive frequency parameters, modeled as learnable variables with context-dependent offsets. 
This design enables adaptive adjustment of the frequency information and enhances the sensitivity to the underlying temporal dynamics. Specifically, for each channel $\mathbf{U}_{d}\in\mathbb{R}^{M\times V}$ in $\mathbf{U}$, we first define the fundamental Fourier frequencies:
 \begin{align}
   \boldsymbol{\omega}_{base}&=\left [ \frac{2\pi k}{V} \right ]_{k=0}^{V-1}\in \mathbb{R}^{V}  .  
 \end{align}

To achieve the context-aware frequency analysis, we employ a global average pooling $\mathrm{AvgPool}(\cdot)$ to aggregate temporal information across all $M$ patches, and a lightweight adapter then generates the offset $\boldsymbol{\Delta\omega}$. The context-aware frequency basis $\boldsymbol{\omega}$ is given by:
 \begin{align}
    \boldsymbol{\mathbf{\Delta\omega}}&=Adapter(Avgpooling(\mathbf{U}_{d}))\in \mathbb{R}^{V}, \\
    \boldsymbol{\omega}&=\boldsymbol{\omega}_{base}+\boldsymbol{\mathbf{\Delta\omega}}, 
 \end{align}
where $Adapter(\cdot)$ comprises two fully connected layers interspersed with an activation function.

\begin{algorithm}[t]
\caption{Adaptive Frequency-Gated State Space Update}
\label{alg:main}
\begin{algorithmic}[1]
\Require $\mathbf{U}\in\mathbb{R}^{D\times M\times V}$
\Ensure $\mathbf{Z}\in\mathbb{R}^{D\times M\times V}$
\For{each channel $\mathbf{U}_{d}\in\mathbb{R}^{M\times V}$ in $\mathbf{U}$}
    \State $\triangleright$ \emph{Adaptive Frequency Computation}
    \State $\bar{\mathbf{U}}=\mathrm{AvgPool}(\mathbf{U}_{d})\in\mathbb{R}^{V}$
    \State $\boldsymbol{\omega}_{base}=\left[2\pi k/V\right]_{k=0}^{V-1}\in\mathbb{R}^{V}$
    \State $\boldsymbol{\Delta\omega}=\mathrm{Linear}(\mathrm{ReLU}(\mathrm{Linear}(\bar{\mathbf{U}})))\in\mathbb{R}^{V}$
    \State $\boldsymbol{\omega}=\boldsymbol{\omega}_{base}+\boldsymbol{\Delta\omega}$
    \State $\triangleright$ \emph{Frequency-gated Hidden States Update}
    \For{$m=1$ to $M$}
        \State $\triangleright$ \emph{Trigonometric Modulation}
        \State $\mathbf{c}_{m}=[\cos\omega_{1}m,\cdots,\cos\omega_{V}m]$
        \State $\mathbf{s}_{m}=[\sin\omega_{1}m,\cdots,\sin\omega_{V}m]$
        \State $\triangleright$ \emph{State Gating}
        \State $\mathbf{A}^{time}_{m}=\sigma(\mathbf{M}^{time}_{u}\mathbf{u}_{m}+\mathbf{M}^{time}_{z}\mathbf{z}_{m-1})\in\mathbb{R}^{V}$
        \State $\mathbf{A}^{fre}_{m}=\sigma(\mathbf{M}^{fre}_{u}\mathbf{u}_{m}+\mathbf{M}^{fre}_{z}\mathbf{z}_{m-1})\in\mathbb{R}^{S}$
        \State $\mathbf{A}_{m}=\mathbf{A}^{fre}_{m}\otimes\mathbf{A}^{time}_{m}$
        \State $\triangleright$ \emph{State Equations}
        \State $\mathbf{f}^{re}_{m}=\mathbf{A}_{m}\odot\mathbf{f}^{re}_{m-1}+\mathbf{B}_{m}\mathbf{c}_{m}$
        \State $\mathbf{f}^{im}_{m}=\mathbf{A}_{m}\odot\mathbf{f}^{im}_{m-1}+\mathbf{B}_{m}\mathbf{s}_{m}$
        \State $\mathbf{E}^{Amp}_{m}=\sqrt{(\mathbf{f}^{re}_{m})^{2}+(\mathbf{f}^{im}_{m})^{2}}$
        \State $\triangleright$ \emph{Output Gating}
        \State $\mathbf{g}_{m}=\sigma(\mathbf{W}^{Amp}_{g}\mathbf{E}^{Amp}_{m}+\mathbf{W}^{u}_{g}\mathbf{u}_{m}+\mathbf{W}^{y}_{g}\mathbf{y}_{m-1})$
        \State $\mathbf{y}_{m}=\mathbf{C}_{m}\mathbf{E}^{Amp}_{m}+\mathbf{D}_{m}^{u}\mathbf{u}_{m}+\mathbf{D}_{m}^{y}\mathbf{y}_{m-1}$
        \State $\mathbf{z}_{m}={\textstyle\sum_{s=0}^{S-1}}\mathbf{g}_{m}\odot\mathbf{y}_{m}\in\mathbb{R}^{V}$
    \EndFor
    \State Stack the results $\mathbf{z}_{1},\dots,\mathbf{z}_{M}$ into $\mathbf{Z}_{d}\in\mathbb{R}^{M\times V}$
\EndFor
\State Stack $\mathbf{Z}_{d}$ across all channels into $\mathbf{Z}$
\State \Return $\mathbf{Z}$
\end{algorithmic}
\end{algorithm}

\subsubsection{Frequency-gated Hidden States Update}
To facilitate a seamless synergy between adaptive frequency modeling and long-range dependency learning in SSMs, we propose a frequency-gated hidden states update mechanism.
Concretely, for each row $\mathbf{u}_{m} \in \mathbb{R}^{1\times V}$ in $\mathbf{U}_{d} \in \mathbb{R}^{M\times V}$, 
we employ the learnable frequency basis $\boldsymbol\omega$ to obtain the frequency-domain representation $\mathbf{u}_{m}^{\omega } \in \mathbb{C}^{1\times V}$, and extend the temporal hidden states into the time–frequency representation $\mathbf{f}_{m} \in \mathbb{C}^{S\times V}$, where the rows and columns of $\mathbf{f}_{m}$ correspond to $S$-dimensional frequencies and $V$-dimensional temporal states, respectively. Since the transition matrix in selective SSMs is typically diagonal, we set $S$$=$$V$ to preserve the training benefits associated with this diagonal structure. 

We further extend the temporal forgetting gate in the original selective SSMs into a unified time-frequency forgetting gate $\mathbf{A}_{m}\in\mathbb{R}^{S\times V}$, enabling simultaneous utilization of temporal and frequency representations for selective filtering of relevant information:
\begin{align}
\mathbf{A}_{m}&=\mathbf{A}^{fre}_{m}\otimes\mathbf{A}^{time}_{m}\in\mathbb{R}^{S\times V},\\
\mathbf{A}^{time}_{m}&=\sigma(\mathbf{M}^{time}_{u}\mathbf{u}_{m}+\mathbf{M}^{time}_{z}\mathbf{z}_{m-1})\in\mathbb{R}^{V},\\
\mathbf{A}^{fre}_{m}&=\sigma(\mathbf{M}^{fre}_{u}\mathbf{u}_{m}+\mathbf{M}^{fre}_{z}\mathbf{z}_{m-1})\in\mathbb{R}^{S},
\label{eq:time_frequency_Am}
\end{align}
where $\otimes$ denotes the outer product, $\mathbf{M}^{time}_{u},\mathbf{M}^{time}_{z}\in\mathbb{R}^{V\times V}$, and $\mathbf{M}^{fre}_{u},\mathbf{M}^{fre}_{z}\in\mathbb{R}^{S\times V}$. Accordingly, the state equation is reformulated as:
\begin{equation}
\label{eq:state_equation}
\mathbf{f}_{m}=\mathbf{A}_{m}\odot\mathbf{f}_{m-1}+\mathbf{B}_{m}\mathbf{u}_{m}^{\omega},
\end{equation}
where $\mathbf{u}_{m}^{\omega}=[e^{j\omega_{1}m},\cdots,e^{j\omega_{V}m}]$, $j=\sqrt{-1}$, and $\mathbf{B}_{m}\in\mathbb{R}^{S\times1}$ controls the incorporation of input-related information into the time-frequency state updates.


To maintain computational simplicity and avoid operations in the complex domain, we update the real and imaginary components of $\mathbf{f}_{m}$ separately. Let $\mathbf{c}_{m}=[\cos\omega_{1}m,\cdots,\cos\omega_{V}m]$ and $\mathbf{s}_{m}=[\sin\omega_{1}m,\cdots,\sin\omega_{V}m]$; then
\begin{align}
\mathbf{f}_{m}^{re}&=\mathbf{A}_{m}\odot\mathbf{f}_{m-1}^{re}+\mathbf{B}_{m}\mathbf{c}_{m}, \quad
\mathbf{f}_{m}^{im}=\mathbf{A}_{m}\odot\mathbf{f}_{m-1}^{im}+\mathbf{B}_{m}\mathbf{s}_{m}.
\label{eq:real_imag}
\end{align}
This decomposition enables real-valued arithmetic while preserving the structural information encoded in the complex frequency representation.

For the output equation, rather than directly using the complex representation $\mathbf{f}_{m}$, we characterize each frequency component by its amplitude and phase, i.e., its magnitude and shift:
\begin{align}
\mathbf{E}^{Amp}_{m}&=\left|\mathbf{f}_{m}\right|=\sqrt{(\mathbf{f}_{m}^{re})^{2}+(\mathbf{f}_{m}^{im})^{2}}\in\mathbb{R}^{S\times V},\\
\mathbf{E}^{Phase}_{m}&=\arctan\left(\frac{\mathbf{f}_{m}^{im}}{\mathbf{f}_{m}^{re}}\right)\in\left[-\frac{\pi}{2},\frac{\pi}{2}\right]^{S\times V}.
\label{eq:amp_phase}
\end{align}
%
%
Based on the theoretical analysis in Appendix~\ref{app:phase_optimization} and the empirical findings in Table~\ref{tab:amp_phase_ablation}, we exclude the phase component $\mathbf{E}^{Phase}_{m}$ and adopt the amplitude component $\mathbf{E}^{Amp}_{m}$ as the primary frequency representation. This preference is motivated by the amplitude's capacity to capture periodic strength and its inherent robustness to temporal shifts. Although phase can encode alignment information, our experiments show that explicit phase modeling does not improve forecasting accuracy in the evaluated LTSF settings and instead introduces additional optimization sensitivity. Therefore, the output expression relies only on $\mathbf{E}^{Amp}_{m}$. Moreover, the inclusion of frequency-domain analysis requires extending the residual term of the original selective SSM. The resulting formulation jointly integrates the time-domain input $\mathbf{u}_{m}\in\mathbb{R}^{1\times V}$ and the previous time-frequency state $\mathbf{y}_{m-1}\in\mathbb{R}^{S\times V}$:
\begin{equation}
\label{eq:output_equation}
\mathbf{y}_{m}=\mathbf{C}_{m}\mathbf{E}^{Amp}_{m}+\mathbf{D}_{m}^{u}\mathbf{u}_{m}+\mathbf{D}_{m}^{y}\mathbf{y}_{m-1},
\end{equation}
where $\mathbf{C}_{m}\in\mathbb{R}^{S\times S}$, $\mathbf{D}_{m}^{u}\in\mathbb{R}^{S\times1}$, and $\mathbf{D}_{m}^{y}\in\mathbb{R}^{S\times S}$.


To derive the final representation, we aggregate the $S$ frequency rows of $\mathbf{y}_{m}\in\mathbb{R}^{S\times V}$. However, direct summation indiscriminately merges all contributions, which may destabilize training and over-smooth discriminative spectral information. 
To overcome this limitation, we introduce a gating mechanism that selectively preserves and emphasizes task-relevant periodic information when producing the final output:

  \begin{align}
    \mathbf{g}_m& =\sigma (\mathbf{W}^{Amp}_{g}{\mathbf{E}}^{Amp}_m+\mathbf{W}^{u}_{g}\mathbf{u}_{m}+\mathbf{W}^{y}_{g}\mathbf{y}_{m-1}),\\
    \mathbf{z}_m& =  {\textstyle \sum_{s=0}^{S-1}}  \mathbf{g}_m \odot \mathbf{y}_m \in \mathbb{R}^{V},
  \label{eq:gated_output_equation}
  \end{align}
where $\mathbf{W}^{Amp}_{g}\in \mathbb{R}^{S\times S} $, $\mathbf{W}^{u}_{g}\in \mathbb{R}^{S\times 1} $, $\mathbf{W}^{y}_{g}\in \mathbb{R}^{S\times S} $, and $\sigma(\cdot)$ is an element-wise activation function.




Finally, we stack the results $\mathbf{Z}_{d}\in\mathbb{R}^{M\times V}$ from all channels of $\mathbf{U}$ to form the frequency-gated temporal representation $\mathbf{Z}\in\mathbb{R}^{D\times M\times V}$. The whole procedure of the adaptive frequency-gated state space module is detailed in Algorithm~\ref{alg:main}.

\subsection{Forecasting Layer}
For final forecasting, we flatten the frequency-gated temporal representation $\mathbf{Z}$ produced by the adaptive frequency-gated state space module and apply a linear projection to produce the forecasting outputs:
\begin{equation}\label{eq:forecasting}
\mathcal{X}^{\mathrm{Output}}=\text{Linear}(\text{Flatten}(\mathbf{Z}))^T,
\end{equation}
where $\text{Flatten}(\cdot)$ flattens $\mathbf{Z}\in\mathbb{R}^{D\times M\times V}$ into $\mathbf{Z}'\in\mathbb{R}^{D\times(M\cdot V)}$, and a linear layer maps the last dimension from $M\cdot V$ to $H$; the result is then transposed to obtain $\mathcal{X}^{\mathrm{Output}}\in\mathbb{R}^{H\times D}$ defined in Eq.~\eqref{equ_1}.

\begin{table*}[t]
\caption{The average results on seven public datasets with input length $T = 96$ and prediction lengths $H \in \{96, 192, 336, 720\}$ (for the ILI dataset, input length $T = 36$ and prediction lengths $H \in \{24, 36, 48, 60\}$). The best results are highlighted in bold.}
\label{tab:main_results}
\centering
\footnotesize
\resizebox{\textwidth}{!}{%
\begin{tabular}{l ccccccccccccccccccc}
\toprule
\textbf{Models} &
\multicolumn{2}{c}{\textbf{\makecell{\M \\ (Ours)}}} &
\multicolumn{2}{c}{\textbf{\makecell{Affirm \\ (2025)}}} &
\multicolumn{2}{c}{\textbf{\makecell{S-Mamba \\ (2025)}}} &
\multicolumn{2}{c}{\textbf{\makecell{STDMamba \\ (2025)}}} &
\multicolumn{2}{c}{\textbf{\makecell{TimeMachine \\ (2024)}}} &
\multicolumn{2}{c}{\textbf{\makecell{FreTS \\ (2023)}}} &
\multicolumn{2}{c}{\textbf{\makecell{PatchTST \\ (2023)}}} &
\multicolumn{2}{c}{\textbf{\makecell{iTransformer \\ (2024)}}} &
\multicolumn{2}{c}{\textbf{\makecell{ModernTCN \\ (2024)}}} \\
\cmidrule(lr){1-1} \cmidrule(lr){2-3} \cmidrule(lr){4-5} \cmidrule(lr){6-7} \cmidrule(lr){8-9} \cmidrule(lr){10-11} \cmidrule(lr){12-13} \cmidrule(lr){14-15} \cmidrule(lr){16-17} \cmidrule(lr){18-19}
\textbf{Metric} & \textbf{MSE} & \textbf{MAE} & \textbf{MSE} & \textbf{MAE} & \textbf{MSE} & \textbf{MAE} & \textbf{MSE} & \textbf{MAE} & \textbf{MSE} & \textbf{MAE} & \textbf{MSE} & \textbf{MAE} & \textbf{MSE} & \textbf{MAE} & \textbf{MSE} & \textbf{MAE} & \textbf{MSE} & \textbf{MAE} \\
\midrule
\textbf{ETTh1}    & \textbf{0.441} & 0.440 & 0.445 & \textbf{0.438} & 0.455 & 0.450 & 0.446 & 0.447 & 0.461 & 0.447 & 0.475 & 0.462 & 0.469 & 0.454 & 0.454 & 0.447 & 0.478 & 0.468 \\
\textbf{ETTh2}    & \textbf{0.371} & \textbf{0.399} & 0.374 & 0.402 & 0.381 & 0.405 & 0.387 & 0.420 & 0.373 & 0.400 & 0.489 & 0.478 & 0.387 & 0.407 & 0.383 & 0.407 & 0.405 & 0.425 \\
\textbf{ETTm1}    & \textbf{0.378} & \textbf{0.396} & 0.393 & 0.402 & 0.398 & 0.405 & 0.421 & 0.422 & 0.388 & 0.398 & 0.402 & 0.411 & 0.387 & 0.400 & 0.407 & 0.410 & 0.389 & 0.405 \\
\textbf{ETTm2}    & \textbf{0.275} & \textbf{0.322} & 0.288 & 0.332 & 0.288 & 0.332 & 0.293 & 0.334 & 0.281 & 0.326 & 0.305 & 0.357 & 0.281 & 0.326 & 0.288 & 0.332 & 0.291 & 0.335 \\
\textbf{ILI}      & \textbf{1.675} & \textbf{0.802} & 2.023 & 0.930 & 2.060 & 0.933 & 1.926 & 0.868 & 1.892 & 0.842 & 2.862 & 1.125 & 2.177 & 0.913 & 2.087 & 0.888 & 2.153 & 0.905 \\
\textbf{Weather}  & \textbf{0.244} & \textbf{0.272} & 0.260 & 0.280 & 0.251 & 0.276 & 0.271 & 0.290 & 0.260 & 0.279 & 0.255 & 0.300 & 0.259 & 0.281 & 0.258 & 0.278 & 0.260 & 0.280 \\
\textbf{Exchange} & \textbf{0.344} & \textbf{0.394} & 0.350 & 0.396 & 0.367 & 0.408 & 0.358 & 0.406 & 0.348 & 0.396 & 0.399 & 0.432 & 0.367 & 0.404 & 0.360 & 0.403 & 0.346 & \textbf{0.394} \\
\bottomrule
\end{tabular}%
}
\end{table*}

\section{Experiments and Results}
We design extensive experiments to answer the following three research questions (RQs) about \M:
\begin{itemize}
  \item \textbf{RQ1 (Overall effectiveness):} Does \M~achieve superior forecasting accuracy over state-of-the-art baselines across public and domain-specific benchmarks?
  \item \textbf{RQ2 (Controlled validation of the core claim and design rationale):} Is time-frequency heterogeneity itself a major failure mode of existing approaches, and does coupling frequency information \emph{directly} with the state transition causally drive the improvement? How does each key design of the adaptive frequency-gated state space module (AFSSM) contribute?
  \item \textbf{RQ3 (Computational efficiency):} Does \M~remain computationally competitive in both training and inference?
\end{itemize}

\subsection{Experiment Settings}
\textbf{Datasets.} We evaluate \M~on eleven benchmarks, including seven public datasets (four ETT variants: ETTh1, ETTh2, ETTm1, and ETTm2; the influenza-like illness (ILI) dataset; the Weather dataset; and the Exchange Rate dataset) and four domain-specific datasets, i.e., the KnowAir-V2 air quality dataset ($\mathrm{O_3}$-bthsa and PM2.5-bthsa) and the NOAA sea surface temperature dataset (NPO-SST and INO-SST). 
Detailed descriptions of these datasets are provided in Appendix~\ref{Main_Dataset_details}.

\textbf{Baselines and Evaluation Metrics.} We conduct a rigorous comparison of \M~against eight representative state-of-the-art baseline methods, i.e., Affirm~\cite{wu2025affirm}, S-Mamba~\cite{wang2025smamba}, STDMamba~\cite{STDMamba}, TimeMachine~\cite{ahamed2024timemachine}, FreTS~\cite{yi2023frets}, PatchTST~\cite{PatchTST}, iTransformer~\cite{iTransformer}, and ModernTCN~\cite{ModernTCN}, where the first four methods are based on Mamba. Notably, Affirm and FreTS are highly competitive frequency-based forecasting approaches that explicitly exploit frequency-domain representations, serving as critical references for evaluating the effectiveness of \M~in modeling frequency-gated temporal patterns. Following standard practice in LTSF, we adopt the mean squared error (MSE) and the mean absolute error (MAE) to quantify predictive performance.

\textbf{Implementation Details.} We conduct all experiments on an NVIDIA GeForce RTX 4090 GPU, with a 60/20/20 train/validation/test split for the ETT datasets and a 70/10/20 split for the other datasets. For baselines, we report their reported best results if the experimental setups match ours; otherwise, we rerun their official code.
\begin{table}[htbp] 
\centering
\caption{The average results on four domain-specific datasets across all prediction horizons. \textbf{Bold} fonts indicate the best results, while \underline{underlined} fonts signify the second-best results.}
\label{tab:overall_comparison_sstair}

\scriptsize 
\setlength{\tabcolsep}{4.8pt} 
\renewcommand{\arraystretch}{1.04} 

\begin{tabular}{l cccccccc} 
\toprule

\multirow{3}{*}{\textbf{Models}} & 
\multicolumn{4}{c}{\textbf{KnowAir}} & 
\multicolumn{4}{c}{\textbf{NOAA}} \\
\cmidrule(lr){2-5} \cmidrule(lr){6-9}

& \multicolumn{2}{c}{$\mathrm{O_3}$-bthsa} & \multicolumn{2}{c}{PM2.5-bthsa} & \multicolumn{2}{c}{NPO-SST} & \multicolumn{2}{c}{INO-SST} \\
\cmidrule(lr){2-3} \cmidrule(lr){4-5} \cmidrule(lr){6-7} \cmidrule(lr){8-9}

& \textbf{MSE} & \textbf{MAE} & \textbf{MSE} & \textbf{MAE} & \textbf{MSE} & \textbf{MAE} & \textbf{MSE} & \textbf{MAE} \\
\midrule

Affirm       & 1031.018 & 24.138 & 1241.497 & 22.630 & 0.432 & 0.424 & 0.482 & 0.517 \\
S-Mamba     & 1017.048 & 24.009 & 1214.146 & 22.372 & \underline{0.429} & \underline{0.423} & \underline{0.449} & 0.507 \\
STDMamba     & 977.506 & 23.574 & 1217.616 & 22.483 & 0.433 & 0.424 & 0.461 & 0.507 \\
TimeMachine  & \underline{967.756} & \underline{23.362} & \underline{1209.124} & \textbf{22.111} & 0.441 & 0.425 & 0.452 & \underline{0.496} \\
\midrule

FreTS        & 1006.537 & 24.011 & 1222.688 & 25.282 & 0.488 & 0.447 & 0.462 & 0.508 \\
iTransformer & 994.821 & 23.667 & 1210.722 & 22.315 & 0.440 & 0.427 & 0.456 & 0.499 \\
ModernTCN    & 1018.678 & 24.034 & 1276.572 & 23.000 & 0.487 & 0.453 & 0.462 & 0.512 \\
PatchTST     & 992.864 & 23.713 & 1223.610 & 22.414 & 0.486 & 0.450 & 0.492 & 0.523 \\
\midrule

\textbf{\M} & \textbf{940.284} & \textbf{23.109} & \textbf{1206.149} & \underline{22.281} & \textbf{0.423} & \textbf{0.419} & \textbf{0.435} & \textbf{0.491} \\
\bottomrule

\end{tabular}
\end{table}

\subsection{RQ1: Overall Forecasting Performance}

Tables~\ref{tab:main_results} and~\ref{tab:overall_comparison_sstair} show that \M~achieves strong and consistent performance across the evaluated datasets. The gains are particularly clear on the average prediction-length metrics, where \M~improves over both Mamba-based baselines and frequency-domain forecasting methods. On the ILI dataset, \M~reduces the MSE from 1.892 (the best competing result) to 1.675, corresponding to an 11.47\% relative reduction. This improvement indicates that the proposed time-frequency state update is effective in settings with complex and irregular temporal patterns.
The comparison with Affirm and FreTS is especially informative because both methods explicitly exploit frequency-domain representations. Although these baselines remain competitive at some horizons, their overall performance is lower than that of \M. 
For air quality forecasting ($T=72, H=72$, following~\cite{wang2025knowair}) and sea surface temperature prediction ($T=360, H \in \{30, 60, 90, 180, 270, 360\}$, following~\cite{STDMamba}), \M~consistently achieves the lowest average MSE. 
This suggests that frequency analysis is most effective when it is coupled with the state transition itself, rather than applied as an external filter or auxiliary transformation. 

\subsection{RQ2: Controlled Validation of the Core Claim and Design Rationality}

The central claim of \M~has two parts: (i) time-frequency heterogeneity is a major failure mode of existing approaches, and (ii) the improvement comes from \emph{directly} letting frequency information modulate the state transition. We validate both under controlled conditions, and further dissect the contribution of each key design.

\subsubsection{Synthetic Benchmarks with Tunable Time-Frequency Heterogeneity}
\begin{figure}[t]
  \centering
  \includegraphics[width=\linewidth]{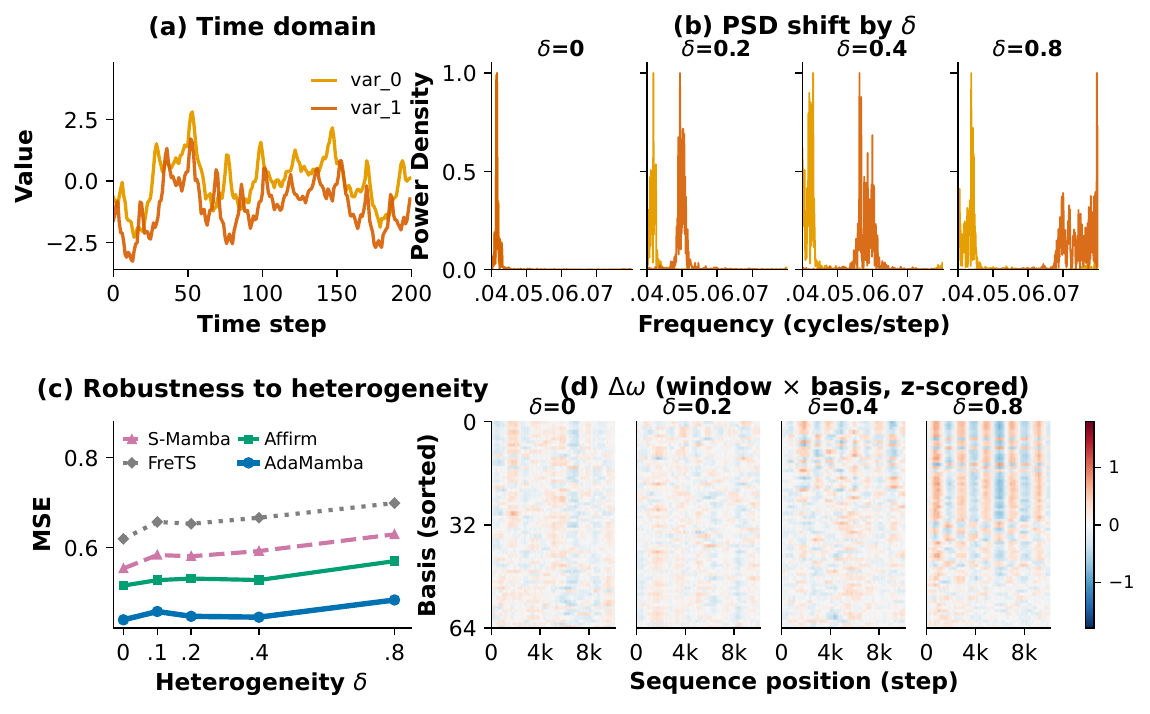}
  \caption{Synthetic benchmarks with tunable time-frequency heterogeneity. (a) A representative variable pair in the time domain, which shares an identical low-frequency envelope and thus exhibits synchronized co-movement. (b) The corresponding spectra: the carrier peak of var$_0$ stays anchored at $f_0$, whereas that of its pair shifts by $\delta\cdot f_0$, with a modulation-induced spread that widens as $\delta$ grows. (c) MSE of \M~and representative baselines versus $\delta$, where $\delta=0$ serves as the time-frequency homogeneous control. (d) Frequency-basis offsets $\Delta\boldsymbol{\omega}$ over sliding windows for each $\delta$.}
  \label{fig:synthetic}
\end{figure}

We construct synthetic multivariate forecasting tasks in which time-frequency heterogeneity is a controlled variable, following the motivating observation in Fig.~\ref{fig_Introduction}. 
Variables are grouped into pairs that share an identical low-frequency envelope in the time domain, whereas their dominant carrier frequencies differ by a controllable discrepancy $\delta \cdot f_0$, with $f_0$ the shared fundamental frequency. 
The instantaneous frequency of each carrier is further modulated by a dataset-level smooth signal $s(t)$ of fixed period whose depth scales linearly with $\delta$, introducing stronger non-stationarity as $\delta$ grows. 
We generate datasets at $\delta \in \{0, 0.2, 0.4, 0.8\}$. 
Fig.~\ref{fig:synthetic}(a) and (b) visualize a representative pair: the two variables co-move with essentially zero lag in the time domain, yet their carrier peaks drift apart with an increasingly visible modulation-induced spread as $\delta$ grows. 
The pairs thus remain synchronized in time but increasingly misaligned in frequency, which is the key property emphasized in our motivation and renders any static, frequency-specific transformation ineffective. 
Dataset details are provided in Appendix~\ref{Synthetic_Dataset_details}.

Fig.~\ref{fig:synthetic}(c) reports MSE versus $\delta$ for \M and representative baselines. 
As heterogeneity increases, the MSE of all baselines grows substantially, confirming that time-frequency heterogeneity itself is a major failure mode of existing approaches. 
\M~degrades much more slowly: from $\delta=0$ to $\delta=0.8$, its MSE increases by only about $10\%$, whereas the baselines increase more steeply, indicating that its advantage stems from handling heterogeneity rather than additional capacity.

Fig.~\ref{fig:synthetic}(d) plots the frequency-basis offsets $\Delta\boldsymbol{\omega}$ over sliding windows, with rows denoting basis. 
At $\delta=0.8$, the modulation-tracking bases show clear vertical striping: their offsets oscillate with a dominant period within $1\%$ of the injected modulation period, indicating that the dynamic frequency mechanism locks onto the injected drift rather than acting as an unconstrained extra transformation. 
As $\delta$ decreases, the structured offset attenuates in a graded manner, showing that the adaptation strength scales with the heterogeneity level. Together with the widening gap in (c), this provides direct evidence that the robustness of \M~arises from its input-adaptive frequency allocation.

\subsubsection{Ablation Study}
To evaluate the contributions of the key designs in \M~and isolate the effect of the frequency--state coupling, we construct four variants as listed below.

\textbf{Case I}: The full \M~that includes both the interaction learning module and the AFSSM.

\textbf{Case II}: The interaction learning module in \M~is replaced with a linear projection, while the AFSSM is kept unchanged.

\textbf{Case III}: The AFSSM in \M~is replaced with the frequency-enhanced Mamba block proposed in~\cite{wu2025affirm}, while the interaction learning module is retained.

\textbf{Case IV}: The interaction learning module in \M~is replaced with a linear projection, and the AFSSM is replaced with a standard Mamba block.




As shown in Table~\ref{tab:ablation_main}, \M~achieves the best performance across all datasets compared with the ablated variants. The comparison between Cases I and II indicates that the interaction learning module contributes to performance improvements. Moreover, the results of Cases II, III, and IV suggest that integrating frequency modeling seamlessly into the state space update is more effective than replacing it with either a standard Mamba block or an externally enhanced frequency module.

\begin{table}[h!]
\centering
\caption{The average MSE across all prediction lengths on seven datasets for different variants of \M. \textbf{AFSSM} denotes the adaptive frequency-gated state space module.}
\label{tab:ablation_main}

\scriptsize 
\setlength{\tabcolsep}{10pt} 
\renewcommand{\arraystretch}{1} 

\begin{tabular}{l cccc} 
\toprule

\textbf{Modules} & \textbf{I} & \textbf{II} & \textbf{III} & \textbf{IV} \\
\midrule

Interaction Learning Module & $\checkmark$ & $\times$   & $\checkmark$ & $\times$   \\
AFSSM  & $\checkmark$ & $\checkmark$ & $\times$   & $\times$     \\
\midrule 

ETTh1    & \textbf{0.441} & 0.447 & 0.453 & 0.505 \\
ETTh2    & \textbf{0.371} & 0.377 & 0.383 & 0.416 \\
ETTm1    & \textbf{0.378} & 0.391 & 0.392 & 0.414 \\
ETTm2    & \textbf{0.275} & 0.279 & 0.283 & 0.291 \\
Weather  & \textbf{0.244} & 0.253 & 0.259 & 0.267 \\
ILI      & \textbf{1.675} & 2.423 & 2.788 & 2.501 \\
Exchange & \textbf{0.344} & 0.354 & 0.362 & 0.393 \\
\bottomrule

\end{tabular}
\end{table}


\subsubsection{Effectiveness of Amplitude and Phase}
\label{Analysis_Phase}
We decompose the complex-valued time-frequency representation $\mathbf{f}_{m}$ into amplitude ($\mathbf{E}^{Amp}_{m}$) and phase ($\mathbf{E}^{Phase}_{m}$) components. 
Amplitude is invariant to global phase shifts and captures the strength and persistence of periodic patterns, which is what LTSF mainly relies on; phase is tied to the temporal reference frame and is sensitive to sequence-dependent offsets and small perturbations. 
Appendix~\ref{app:phase_optimization} analyzes this sensitivity in detail. 

We therefore compare $\mathbf{E}^{Amp}_{m}$-only, $\mathbf{E}^{Amp}_{m}+\mathbf{E}^{Phase}_{m}$, and $\mathbf{E}^{Phase}_{m}$-only to examine whether the phase component provides useful information. 
Table~\ref{tab:amp_phase_ablation} shows a consistent ordering, $\mathrm{Amp}$-only $<$ $\mathrm{Amp}+\mathrm{Phase}$ $<$ $\mathrm{Phase}$-only, on all four ETT datasets, with amplitude-only achieving the lowest average MSE.
These results support our amplitude-centric design for LTSF.



\begin{table}[t]
\centering
\caption{Effectiveness analysis of amplitude and phase modeling, evaluated using the average MSE across all prediction lengths.}
\label{tab:amp_phase_ablation}
\footnotesize
\setlength{\tabcolsep}{6pt}
\begin{tabular}{l cccc}
\toprule
\textbf{Variant} & \textbf{ETTh1} & \textbf{ETTh2} & \textbf{ETTm1} & \textbf{ETTm2} \\
\midrule
$\mathbf{E}^{Amp}_{m}$-only (Ours) & \textbf{0.441} & \textbf{0.371} & \textbf{0.378} & \textbf{0.275} \\
$\mathbf{E}^{Amp}_{m}+\mathbf{E}^{Phase}_{m}$ & 0.448 & 0.376 & 0.388 & 0.277 \\
$\mathbf{E}^{Phase}_{m}$-only & 0.454 & 0.381 & 0.392 & 0.280 \\
\bottomrule
\end{tabular}
\end{table}

\subsubsection{Effectiveness of Dynamic Frequency Adaptation}
To analyze the effectiveness of dynamic frequency adaptation, we compare the proposed AFSSM with a controlled variant in which the dynamic frequency bases are replaced with a static learnable frequency basis while all other components remain unchanged. As shown in Table~\ref{tab:frequency_dynamic}, the dynamic $\boldsymbol{\omega}$ yields lower average MSE than the static variant on all four datasets. The gains are modest but consistent, providing direct evidence that conditioning the frequency basis on each input is preferable to sharing a single static basis in these settings.

\begin{table}[t]
\centering
\caption{Effectiveness analysis of the dynamic frequency adaptation, evaluated using the average MSE across all prediction lengths.}
\label{tab:frequency_dynamic}
\footnotesize
\setlength{\tabcolsep}{6pt}
\begin{tabular}{l cccc}
\toprule
\textbf{Variant} & \textbf{ETTh2} & \textbf{ETTm2} & \textbf{Weather} & \textbf{Exchange} \\
\midrule
Dynamic $\boldsymbol{\omega}$ (Ours) & \textbf{0.371} & \textbf{0.275} & \textbf{0.244} & \textbf{0.344} \\
Fixed $\boldsymbol{\omega}$ & 0.374 & 0.279 & 0.265 & 0.379 \\
\bottomrule
\end{tabular}
\end{table}

\subsection{RQ3: Computational Efficiency}

\subsubsection{Theoretical Complexity}
The dominant computational overhead of \M~arises from the adaptive frequency-gated state space module, while the interaction learning module introduces only linear overhead, as it merely consists of a Conv1D operation followed by a residual combination. Given an encoded representation of shape $(D,M,V)$, we provide a step-by-step time complexity analysis directly mapped to the pseudocode of Algorithm~\ref{alg:main}:
\begin{itemize}
    \item \textbf{Adaptive Frequency Computation}: Average pooling over the sequence requires $\mathcal{O}(M\cdot V)$ operations, and the subsequent Adapter generates the offset $\boldsymbol{\Delta\omega}\in\mathbb{R}^{V}$, requiring $\mathcal{O}(V^{2})$ operations. \textbf{Complexity}: $\mathcal{O}(M\cdot V+V^{2})$.
    \item \textbf{Trigonometric Modulation and Gating}: For each of the $M$ patches, generating the modulation vectors $\mathbf{c}_{m}$ and $\mathbf{s}_{m}$ takes $\mathcal{O}(V)$. In contrast, computing the gates $\mathbf{A}^{time}_{m}$, $\mathbf{A}^{fre}_{m}$, and $\mathbf{g}_{m}$ involves dense matrix projections; for example, $\mathbf{W}^{Amp}_{g}\mathbf{E}^{Amp}_{m}$ projects an $S\times V$ tensor through an $S\times S$ weight matrix, requiring $\mathcal{O}(S^{2}\cdot V)$ per step. \textbf{Complexity}: $\mathcal{O}(M\cdot S^{2}\cdot V)$.
    \item \textbf{State Equations}: Updating $\mathbf{f}^{re}_{m}$ and $\mathbf{f}^{im}_{m}$ involves element-wise Hadamard products ($\mathbf{A}_{m}\odot\mathbf{f}_{m-1}$) and broadcast additions over the $S\times V$ dimension, and computing the amplitude $\mathbf{E}^{Amp}_{m}$ likewise requires element-wise operations on $S\times V$ matrices. \textbf{Complexity}: $\mathcal{O}(M\cdot S\cdot V)$.
    \item \textbf{Output Equations}: Computing $\mathbf{y}_{m}$ involves matrix multiplications such as $\mathbf{C}_{m}\mathbf{E}^{Amp}_{m}$, requiring $\mathcal{O}(S^{2}\cdot V)$ operations per step. The computation of $\mathbf{z}_{m}\in\mathbb{R}^{V}$ performs a Hadamard product between the $S\times V$ tensors $\mathbf{g}_{m}$ and $\mathbf{y}_{m}$, followed by a summation reduction over the $S$ dimension ($\sum_{s=1}^{S}$), which takes $\mathcal{O}(S\cdot V)$ operations per step. \textbf{Complexity}: $\mathcal{O}(M\cdot(S^{2}\cdot V+S\cdot V))$.
\end{itemize}
In sum, the overall complexity is $\mathcal{O}(2MS^{2}V+2MSV+MV+V^{2})$, which simplifies to $\mathcal{O}(MS^{2}V+V^{2})$ with the $MS^{2}V$ term dominating. Since $M=K\times Q$ grows linearly with the look-back length $T$, \M~scales linearly with the sequence length and thus avoids the quadratic cost of Transformer-based architectures, enabling practical computational efficiency in LTSF.



\subsubsection{Runtime Overhead}
\begin{figure}[h!]
  \centering
  \includegraphics[width=0.9\linewidth]{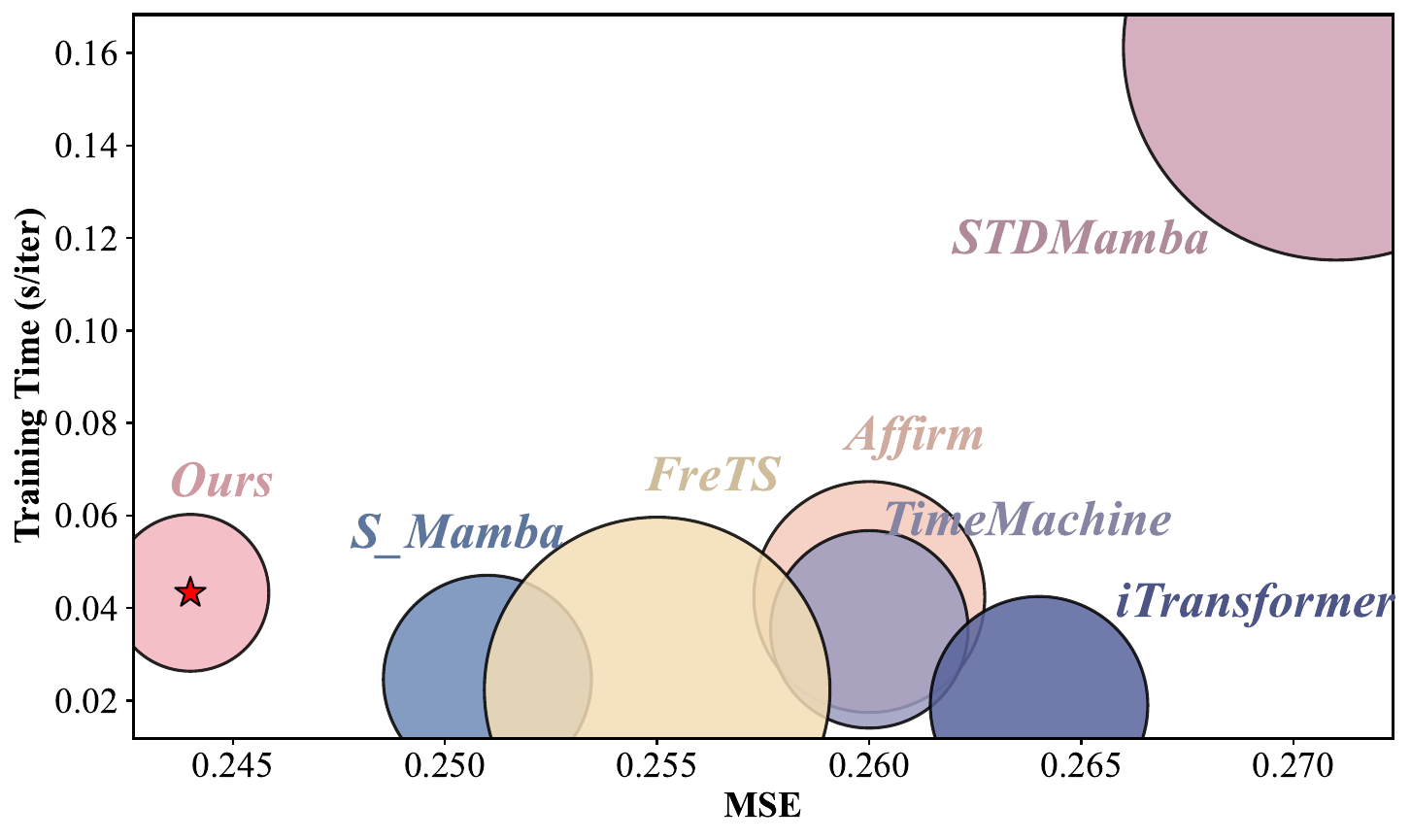}
  \caption{Comparison of \M~and six baseline models on MSE, training time, and parameter counts for 720-step forecasting on the Weather dataset.}
  \label{fig:efficiency}
\end{figure}

All measurements use the Weather dataset with a 96-step input and a 720-step forecasting, batch size 24, on a single NVIDIA RTX 4090. 
Fig.~\ref{fig:efficiency} compares training
cost and accuracy against six baselines, where the horizontal axis is MSE, the
vertical axis is training time per iteration, and bubble size is the parameter
count. 
Table~\ref{tab:rq3-hwfree} further compares the inference costs of \M~with those of four Mamba-based forecasters.


As shown in Fig.~\ref{fig:efficiency}, \M~achieves low MSE with training time comparable to the fastest baselines and a smaller parameter footprint than S-Mamba and STDMamba. 
Table~\ref{tab:rq3-hwfree} places \M~in the middle of the inference trade-off: its MAC count is lower than S-Mamba and STDMamba but higher than TimeMachine and Affirm, its latency is lower than STDMamba and TimeMachine but higher than S-Mamba and Affirm, and its memory is higher than the three compact baselines but lower than TimeMachine. 
Overall, \M~remains competitive, balancing predictive accuracy and computational cost.

\begin{table}[h!]
\centering
\caption{Inference cost of \M~and four representative Mamba-based forecasters for 720-step forecasting on the Weather dataset, measured by the MACs per batch, the peak GPU memory and the latency per batch.}
\label{tab:rq3-hwfree}
\footnotesize
\setlength{\tabcolsep}{2pt}
  \begin{tabular}{lccccc}
  \toprule
  Model & MACs/batch (G) & Peak Mem. (GB) & Latency (ms/batch) \\
\midrule
Affirm         & 0.15 & 0.14 & 5.98  \\
S-Mamba        & 3.51 & 0.05 & 3.46  \\
STDMamba       & 7.65 & 0.10 & 30.62 \\
TimeMachine    & 1.94 & 1.07 & 48.16 \\
AdaMamba (Ours)& 2.29 & 0.47 & 8.75  \\
  \bottomrule
\end{tabular}
\end{table}

\section{Conclusion}

In this work, we propose \M~to address time-frequency heterogeneity in long-term time series forecasting. 
\M~combines an interaction learning
module for cross-variable and multi-scale temporal dynamics with an adaptive frequency-gated state space module that generates input-dependent frequency bases and uses a unified time-frequency gate to directly modulate state evolution.
Across seven public benchmarks and four domain-specific datasets, \M~improves forecasting accuracy at competitive computational cost. Future work will
explore more efficient frequency-state interactions and extend \M~to
large-scale pre-training and spatio-temporal forecasting.

\bibliographystyle{ACM-Reference-Format}
\bibliography{sample-base}

@article{lim2021time,
  title={Time-series forecasting with deep learning: a survey},
  author={Lim, Bryan and Zohren, Stefan},
  journal={Philosophical transactions of the royal society a: mathematical, physical and engineering sciences},
  volume={379},
  number={2194},
  year={2021},
  publisher={The Royal Society}
}

@inproceedings{wen2022robust,
  title={Robust Time Series Analysis and Applications: An Industrial Perspective.},
  author={Wen, Qingsong and Yang, Linxiao and Zhou, Tian and Sun, Liang},
  booktitle={KDD},
  volume={22},
  pages={4836--4837},
  year={2022}
}

@article{liu2022scinet,
  title={Scinet: Time series modeling and forecasting with sample convolution and interaction},
  author={Liu, Minhao and Zeng, Ailing and Chen, Muxi and Xu, Zhijian and Lai, Qiuxia and Ma, Lingna and Xu, Qiang},
  journal={Advances in neural information processing systems},
  volume={35},
  pages={5816--5828},
  year={2022}
}

@inproceedings{zhang2015daily,
  title={Daily-aware personalized recommendation based on feature-level time series analysis},
  author={Zhang, Yongfeng and Zhang, Min and Zhang, Yi and Lai, Guokun and Liu, Yiqun and Zhang, Honghui and Ma, Shaoping},
  booktitle={Proceedings of the 24th international conference on world wide web},
  pages={1373--1383},
  year={2015}
}

@article{li2026asgmamba,
  title={ASGMamba: Adaptive spectral gating Mamba for multivariate time series forecasting},
  author={Li, Qianyang and Zhang, Xingjun and Wang, Shaoxun and Wei, Jia},
  journal={Neurocomputing},
  pages={135084},
  year={2026},
  publisher={Elsevier}
}

@article{barigozzi2019nets,
  title={Nets: Network estimation for time series},
  author={Barigozzi, Matteo and Brownlees, Christian},
  journal={Journal of Applied Econometrics},
  volume={34},
  number={3},
  pages={347--364},
  year={2019},
  publisher={Wiley Online Library}
}

@ARTICLE{STDMamba,
  author={Jiang, Xudong and Wang, Shuyu and Li, Wengen and Yang, Hanchen and Guan, Jihong and Zhang, Yichao and Zhou, Shuigeng},
  journal={IEEE Transactions on Geoscience and Remote Sensing}, 
  title={STDMamba: Spatiotemporal Decomposition Mamba for Long-Term Fine-Grained SST Prediction}, 
  year={2025},
  volume={63},
  number={},
  pages={1-16},
  doi={10.1109/TGRS.2025.3624051}}

@misc{Mamba,
      title={Mamba: Linear-Time Sequence Modeling with Selective State Spaces}, 
      author={Albert Gu and Tri Dao},
      year={2024},
      eprint={2312.00752},
      archivePrefix={arXiv},
      primaryClass={cs.LG},
      url={https://arxiv.org/abs/2312.00752}, 
}

@inproceedings{wu2025affirm,
  title={Affirm: Interactive mamba with adaptive fourier filters for long-term time series forecasting},
  author={Wu, Yuhan and Meng, Xiyu and Hu, Huajin and Zhang, Junru and Dong, Yabo and Lu, Dongming},
  booktitle={Proceedings of the AAAI Conference on Artificial Intelligence},
  volume={39},
  number={20},
  pages={21599--21607},
  year={2025}
}

@incollection{ahamed2024timemachine,
  title={Timemachine: A time series is worth 4 mambas for long-term forecasting},
  author={Ahamed, Md Atik and Cheng, Qiang},
  booktitle={ECAI 2024},
  pages={1688--1695},
  year={2024},
  publisher={IOS Press}
}

@article{wang2025smamba,
  title={Is mamba effective for time series forecasting?},
  author={Wang, Zihan and Kong, Fanheng and Feng, Shi and Wang, Ming and Yang, Xiaocui and Zhao, Han and Wang, Daling and Zhang, Yifei},
  journal={Neurocomputing},
  volume={619},
  pages={129178},
  year={2025},
  publisher={Elsevier}
}

@article{wu2021autoformer,
  title={Autoformer: Decomposition transformers with auto-correlation for long-term series forecasting},
  author={Wu, Haixu and Xu, Jiehui and Wang, Jianmin and Long, Mingsheng},
  journal={Advances in neural information processing systems},
  volume={34},
  pages={22419--22430},
  year={2021}
}

@inproceedings{zhou2022fedformer,
  title={Fedformer: Frequency enhanced decomposed transformer for long-term series forecasting},
  author={Zhou, Tian and Ma, Ziqing and Wen, Qingsong and Wang, Xue and Sun, Liang and Jin, Rong},
  booktitle={International conference on machine learning},
  pages={27268--27286},
  year={2022},
  organization={PMLR}
}

@article{wang2024fredf,
  title={Fredf: Learning to forecast in the frequency domain},
  author={Wang, Hao and Pan, Licheng and Chen, Zhichao and Yang, Degui and Zhang, Sen and Yang, Yifei and Liu, Xinggao and Li, Haoxuan and Tao, Dacheng},
  journal={arXiv preprint arXiv:2402.02399},
  year={2024}
}

@article{yi2023FreTS,
  title={Frequency-domain MLPs are more effective learners in time series forecasting},
  author={Yi, Kun and Zhang, Qi and Fan, Wei and Wang, Shoujin and Wang, Pengyang and He, Hui and An, Ning and Lian, Defu and Cao, Longbing and Niu, Zhendong},
  journal={Advances in Neural Information Processing Systems},
  volume={36},
  pages={76656--76679},
  year={2023}
}

@article{elvander2020defining,
  title={Defining fundamental frequency for almost harmonic signals},
  author={Elvander, Filip and Jakobsson, Andreas},
  journal={IEEE Transactions on Signal Processing},
  volume={68},
  pages={6453--6466},
  year={2020},
  publisher={IEEE}
}

@inproceedings{PatchTST,
title={A Time Series is Worth 64 Words:  Long-term Forecasting with Transformers},
author={Yuqi Nie and Nam H Nguyen and Phanwadee Sinthong and Jayant Kalagnanam},
booktitle={The Eleventh International Conference on Learning Representations },
year={2023},
}

@inproceedings{iTransformer,
title={iTransformer: Inverted Transformers Are Effective for Time Series Forecasting},
author={Yong Liu and Tengge Hu and Haoran Zhang and Haixu Wu and Shiyu Wang and Lintao Ma and Mingsheng Long},
booktitle={The Twelfth International Conference on Learning Representations},
year={2024}
}

@inproceedings{ModernTCN,
  title={Moderntcn: A modern pure convolution structure for general time series analysis},
  author={Luo, Donghao and Wang, Xue},
  booktitle={The twelfth international conference on learning representations},
  pages={1--43},
  year={2024}
}

@inproceedings{zhou2021informer,
  title={Informer: Beyond efficient transformer for long sequence time-series forecasting},
  author={Zhou, Haoyi and Zhang, Shanghang and Peng, Jieqi and Zhang, Shuai and Li, Jianxin and Xiong, Hui and Zhang, Wancai},
  booktitle={Proceedings of the AAAI conference on artificial intelligence},
  volume={35},
  number={12},
  pages={11106--11115},
  year={2021}
}

@article{wang2025knowair,
  title={KnowAir-V2: A Benchmark Dataset for Air Quality Forecasting with PCDCNet},
  author={Wang, Shuo and Cheng, Yun and Meng, Qingye and Saukh, Olga and Zhang, Jiang and Fan, Jingfang and Zhang, Yuanting and Yuan, Xingyuan and Thiele, Lothar},
  year={2025}
}

@inproceedings{zhang2024frnet,
  title={Frnet: Frequency-based rotation network for long-term time series forecasting},
  author={Zhang, Xinyu and Feng, Shanshan and Ma, Jianghong and Lin, Huiwei and Li, Xutao and Ye, Yunming and Li, Fan and Ong, Yew Soon},
  booktitle={Proceedings of the 30th ACM SIGKDD Conference on Knowledge Discovery and Data Mining},
  pages={3586--3597},
  year={2024}
}

@inproceedings{DLinear,
author = {Zeng, Ailing and Chen, Muxi and Zhang, Lei and Xu, Qiang},
title = {Are transformers effective for time series forecasting?},
year = {2023},
isbn = {978-1-57735-880-0},
publisher = {AAAI Press},
url = {https://doi.org/10.1609/aaai.v37i9.26317},
doi = {10.1609/aaai.v37i9.26317},
booktitle = {Proceedings of the Thirty-Seventh AAAI Conference on Artificial Intelligence and Thirty-Fifth Conference on Innovative Applications of Artificial Intelligence and Thirteenth Symposium on Educational Advances in Artificial Intelligence},
articleno = {1248},
numpages = {8},
series = {AAAI'23/IAAI'23/EAAI'23}
}

@article{huang2025timekan,
  title={Timekan: Kan-based frequency decomposition learning architecture for long-term time series forecasting},
  author={Huang, Songtao and Zhao, Zhen and Li, Can and Bai, Lei},
  journal={arXiv preprint arXiv:2502.06910},
  year={2025}
}

@inproceedings{zhou2024sdformer,
  title={Sdformer: transformer with spectral filter and dynamic attention for multivariate time series long-term forecasting},
  author={Zhou, Ziyu and Lyu, Gengyu and Huang, Yiming and Wang, Zihao and Jia, Ziyu and Yang, Zhen},
  booktitle={Proceedings of the Thirty-Third International Joint Conference on Artificial Intelligence (IJCAI-24), Jeju, Republic of Korea},
  pages={3--9},
  year={2024}
}

@article{wu2022timesnet,
  title={Timesnet: Temporal 2d-variation modeling for general time series analysis},
  author={Wu, Haixu and Hu, Tengge and Liu, Yong and Zhou, Hang and Wang, Jianmin and Long, Mingsheng},
  journal={arXiv preprint arXiv:2210.02186},
  year={2022}
}

@inproceedings{xu2024fits,
  title={FITS: Modeling time series with $10 k $ parameters},
  author={Xu, Zhijian and Zeng, Ailing and Xu, Qiang},
  booktitle={International Conference on Learning Representations},
  volume={2024},
  pages={26295--26318},
  year={2024}
}

@article{li2025faith,
  title={FAITH: Frequency-domain Attention In Two Horizons for time series forecasting},
  author={Li, Ruiqi and Jiang, Maowei and Liu, Quangao and Wang, Kai and Feng, Kaiduo and Sun, Yue and Zhou, Xiufang},
  journal={Knowledge-Based Systems},
  volume={309},
  pages={112790},
  year={2025},
  publisher={Elsevier}
}

@inproceedings{ma2026mixlinear,
  title={MixLinear: Extreme Low Resource Multivariate Time Series Forecasting with $0.1 K $ Parameters},
  author={Ma, Aitian and Luo, Dongsheng and Sha, Mo},
  booktitle={International Conference on Learning Representations},
  volume={2026},
  pages={77764--77789},
  year={2026}
}

@article{zhang2025fldmamba,
  title={Fldmamba: Integrating fourier and laplace transform decomposition with mamba for enhanced time series prediction},
  author={Zhang, Qianru and Yu, Chenglei and Wang, Haixin and Yan, Yudong and Cao, Yuansheng and Yiu, Siu-Ming and Wu, Tailin and Yin, Hongzhi},
  journal={arXiv preprint arXiv:2507.12803},
  year={2025}
}

@article{guan2026slan,
  title={SLAN: A State-space Linear Attention Network with Meta Guidance and Chunk-wise Fusion for Long-term Time Series Forecasting},
  author={Guan, Jihong and Jiang, Xudong and Yang, Hanchen and Li, Wengen and Zhang, Yichao and Zhou, Shuigeng and others},
  journal={Expert Systems with Applications},
  pages={133367},
  year={2026},
  publisher={Elsevier}
}

@inproceedings{wang2025csformer,
  title={CSformer: Combining Channel Independence and Mixing for Robust Multivariate Time Series Forecasting},
  author={Wang, Haoxin and Mo, Yipeng and Xiang, Kunlan and Yin, Nan and Dai, Honghe and Li, Bixiong and Fan, Songhai},
  booktitle={Proceedings of the AAAI Conference on Artificial Intelligence},
  volume={39},
  number={20},
  pages={21090--21098},
  year={2025}
}

@article{yang2025frwkv,
  title={FRWKV: Frequency-Domain Linear Attention for Long-Term Time Series Forecasting},
  author={Yang, Qingyuan and Deng, Shizhuo and Chen, Dongyue and Teng, Da and Gan, Zehua},
  journal={arXiv preprint arXiv:2512.07539},
  year={2025}
}

@article{baig2026freqlite,
  title={FreqLite: A Lightweight Frequency-Decomposed Linear Model with Adaptive Reversible Normalization for Robust Long-Term Time-Series Forecasting},
  author={Baig, Mirza Samad Ahmed and Gillani, Syeda Anshrah},
  journal={arXiv preprint arXiv:2606.01339},
  year={2026}
}

@article{zhou2022film,
  title={Film: Frequency improved legendre memory model for long-term time series forecasting},
  author={Zhou, Tian and Ma, Ziqing and Wen, Qingsong and Sun, Liang and Yao, Tao and Yin, Wotao and Jin, Rong and others},
  journal={Advances in neural information processing systems},
  volume={35},
  pages={12677--12690},
  year={2022}
}

@article{zhang2026buttermamba,
  title={ButterMamba: Butterworth-Enhanced Spatial-Temporal Mamba for Efficient Traffic Flow Prediction},
  author={Zhang, Limiao and Lu, Yuhui and Gao, Jie and Jiang, Hao and Ma, Haiping and Zhang, Xingyi},
  journal={arXiv preprint arXiv:2608.29658},
  year={2026}
}

@article{yi2024filternet,
  title={Filternet: Harnessing frequency filters for time series forecasting},
  author={Yi, Kun and Fei, Jingru and Zhang, Qi and He, Hui and Hao, Shufeng and Lian, Defu and Fan, Wei},
  journal={Advances in neural information processing systems},
  volume={37},
  pages={55115--55140},
  year={2024}
}

@inproceedings{dayag2026filter,
  title={Filter then attend: improving attention-based time series forecasting with spectral filtering},
  author={Dayag, Elisha and Tran, Nhat Thanh and Xin, Jack},
  booktitle={ICASSP 2026-2026 IEEE International Conference on Acoustics, Speech and Signal Processing (ICASSP)},
  pages={1131--1135},
  year={2026},
  organization={IEEE}
}

@article{chen2026freqlens,
  title={FreqLens: Interpretable Frequency Attribution for Time Series Forecasting},
  author={Chen, Chi-Sheng and Zhang, Xinyu and Kuo, En-Jui and Chen, Guan-Ying and Xie, Qiuzhe and Zhang, Fan},
  journal={arXiv preprint arXiv:2602.08768},
  year={2026}
}

@inproceedings{wang2025filterts,
  title={Filterts: Comprehensive frequency filtering for multivariate time series forecasting},
  author={Wang, Yulong and Liu, Yushuo and Duan, Xiaoyi and Wang, Kai},
  booktitle={Proceedings of the AAAI Conference on Artificial Intelligence},
  volume={39},
  number={20},
  pages={21375--21383},
  year={2025}
}

@article{wu2026kfs,
  title={KFS: KAN based adaptive frequency selection learning architecture for long-term time series forecasting},
  author={Wu, Changning and Wu, Gao and Cai, Rongyao and Liu, Yong and Zhang, Kexin},
  journal={Advanced Engineering Informatics},
  volume={72},
  pages={104442},
  year={2026},
  publisher={Elsevier}
}



\appendix

\section{Dataset Details.}
\subsection{Public benchmarks and domain-specific datasets}\label{Main_Dataset_details}
We conduct extensive experiments on seven public LTSF benchmarks and four domain-specific datasets covering multiple domains and temporal characteristics. The statistical characteristics of these datasets are summarized in Table~\ref{tab:dataset}: 

(1) ETT~\cite{zhou2021informer} consists of two subsets, ETTh and ETTm, which record power transformer temperature measurements from July 2016 to July 2018 at sampling intervals of 1 hour and 15 minutes, respectively.

(2) Weather~\cite{zhou2021informer} provides complete observations of 21 meteorological variables recorded at 10-minute intervals throughout the year 2020.

(3) ILI~\cite{wu2021autoformer} includes the weekly recorded influenza-like illness (ILI) patients data from Centers for Disease Control and Prevention of the United States between 2002 and 2021,which describes the ratio of patients seen with ILI and the total number of the patients. 

(4) Exchange~\cite{wu2021autoformer} records the daily exchange rates of eight different countries ranging from 1990 to 2016.

(5) KnowAir-V2~\cite{wang2025knowair} has one major densely populated region in China: the Beijing–Tianjin–Hebei and Surrounding Areas (BTHSA), comprising data from 228 monitoring stations. The dataset spans from January 1, 2016, to December 31, 2023, providing an extensive temporal coverage for model training and evaluation.


(6) NOAA-SST~\cite{STDMamba} includes two study areas, that is, Northwest Pacific Ocean ([1.0°N-49.0°N,101.0°E-179.0°E]) and Indian Ocean ([49.0°S-5.0°N,41.0°E-105.0°E]). The SST data are from the daily average data in the Optimum Interpolation High-Resolution SST Dataset Version 2 (OISST), provided by the Physical Sciences Laboratory of the National Oceanic and Atmospheric Administration (NOAA). 
Given a spatial resolution of 2° × 2°, the Northwest Pacific Ocean comprises 25 grids along the latitude and 40 grids along the longitude, resulting in a total of 1,000 grid regions. 
Similarly, the Indian Ocean consists of 28 grids along the latitude and 33 grids along the longitude, yielding a total of 924 grid regions. 
These two datasets span from January 2005 to the present and are referred to as NPO-SST and INO-SST, respectively.

\begin{table}[htbp]
\caption{Statistics of datasets in our experiments.}
\centering
\begin{tabular}{l|ccc} 
\toprule
Dataset     & Sequence Length & Features & Frequency \\ 
\midrule
ETTh1       & 17,420          & 7        & 1h        \\
ETTh2       & 17,420          & 7        & 1h        \\
ETTm1       & 69,680          & 7        & 15 min    \\
ETTm2       & 69,680          & 7        & 15 min    \\
Weather     & 52,696          & 21       & 10 min    \\
ILI     & 966          & 7      & 7d        \\
Exchange & 7,587          & 8      & 1d        \\
KnowAir-{$\mathrm{O_3}$-bthsa}    & 70,128     & 228      & 1h \\
KnowAir-{$\mathrm{PM2.5}$-bthsa}    & 70,128     & 228      & 1h \\
NPO-SST     & 7,549          & 1000     & 1d \\
INO-SST     & 7,549          & 924      & 1d \\
\bottomrule
\end{tabular}

\label{tab:dataset}
\end{table}

\subsection{Synthetic Benchmarks with Tunable Time-Frequency Heterogeneity}
\label{Synthetic_Dataset_details}

\paragraph{Design goals.}
The construction has three goals: paired variables are zero-lag synchronized in
time yet separated in frequency by a controlled amount; the separation cannot be
absorbed by any fixed linear recurrence, so a static frequency allocation is
insufficient; and a single scalar $\delta$ indexes the difficulty while all else
is held fixed.

\paragraph{Construction.}
Each dataset contains $V=20$ variables in $P=10$ pairs, with members $x_{2k}(t)$
and $x_{2k+1}(t)$. All stochastic quantities are drawn once from a fixed seed and
reused across $\delta$, so datasets differ only in $\delta$.

Within a pair, all non-carrier components are shared, rendering the two variables
zero-lag synchronized in time:
\begin{align}
e_k(t) &= 1+\tfrac{1}{2}\sin\!\big(2\pi t/T_{\mathrm{env}}+\phi_k\big),
  \qquad T_{\mathrm{env}}=720, \label{eq:syn-env}\\
g(t) &= 0.1+0.9\,\hat g(t),\qquad
  \hat g(t)=\mathrm{minmax}\big(\mathrm{MA}_{24}(\xi)\big),\quad
  \xi_t\stackrel{\mathrm{iid}}{\sim}\mathcal{N}(0,1), \label{eq:syn-gate}\\
c_k(t) &= \big(\bar c_k(t)-\mathbb{E}[\bar c_k]\big)\big/\mathrm{std}(\bar c_k),\qquad
  \bar c_k(t)=\sum_{i}\theta_{k,i}\sin\!\big(2\pi t/P_{k,i}+\varphi_{k,i}\big),
  \label{eq:syn-comove}\\
\psi_k(t) &= \psi_{k,0}+\sum_{s\le t}\eta_{k,s},\qquad
  \eta_{k,s}\stackrel{\mathrm{iid}}{\sim}\mathcal{N}(0,\sigma_\phi^2),\quad
  \sigma_\phi=0.08, \label{eq:syn-walk}
\end{align}
Here $e_k$ is a slow envelope, $g$ a burst gate, $c_k$ a standardized zero-lag
multi-sine (periods alternating between $\{96,72,56,144\}$ and
$\{192,432,288\}$ steps), and $\psi_k$ a random-walk phase. The gate $g$ is a
single global realization shared by all variables, representing burst timing
visible across channels; since it is shared within pairs either way, a per-pair
gate would only change cross-pair statistics. All shared periods lie below
$f_{\mathrm{low}}=0.02$ cycles/step ($1/56\approx0.0179$), outside the carrier
band.

The two carriers share modulation, walk, and weights, differing only in center
frequency:
\begin{equation}
u_{k,j}(t)=\sum_{m=1}^{M} w_{k,m}\,
\cos\!\Big(2\pi m\big(F_{k,j}\,t+F_{k,j}\,\varepsilon S(t)\big)+m\,\psi_k(t)\Big),
\qquad M=5, \label{eq:syn-carrier}
\end{equation}
with $F_{k,0}=f_0$, $F_{k,1}=(1+\delta)f_0$, $f_0=1/24$, modulation depth
$\varepsilon=\beta\delta$, $\beta=0.1$, and $s(t)=\sin(2\pi t/T_{\mathrm{fm}}+\chi)$,
$S(t)=-(T_{\mathrm{fm}}/2\pi)\cos(2\pi t/T_{\mathrm{fm}}+\chi)$,
$T_{\mathrm{fm}}=2048$, evaluated in closed form. The observed series is
\begin{equation}
x_{k,j}(t)=A_k\Big[\gamma\,c_k(t)+e_k(t)\,g(t)\,\alpha\,u_{k,j}(t)\Big]
+\sigma\,\zeta_{k,j,t},
 \label{eq:syn-signal}
\end{equation}
where $\gamma=0.8,\ \alpha=1.6,\ \sigma=0.1$, $A_k\sim\mathcal{U}(0.5,2)$ and $\zeta_{k,j,t}\stackrel{\mathrm{iid}}{\sim}
\mathcal{N}(0,1)$ independently per variable and step. The carrier and co-movement
are kept at the same order of magnitude
($\mathrm{std}(e g\alpha u)/\mathrm{std}(\gamma c)=0.49$--$0.63$, mean $0.55$), so
neither dominates by construction.

The weights are $w_{k,1}=1$ and $w_{k,m}=0.45\,r_{k,m}^{\,m-1}$ with
$r_{k,m}\sim\mathcal{U}(0.3,1)$ for $m=2,\dots,5$, normalized to unit sum; thus
$w_{k,1}\ge1/(1+4\times0.45)=0.357$ by construction and $w_{k,1}\in[0.472,0.657]$
for the realized draw, so the fundamental dominates. The highest component at the
largest depth is $M(1+\delta)f_0=5\times1.8/24=0.375<1/2$, so no component
aliases.

The instantaneous frequency of $u_{k,j}$ is
\begin{equation}
\begin{aligned}
    f_{k,j}(t)&=\frac{1}{2\pi}\frac{d}{dt}\Big[2\pi F_{k,j}t+2\pi F_{k,j}\varepsilon
S(t)+\psi_k(t)\Big]\\&=F_{k,j}\big(1+\varepsilon s(t)\big)+\frac{\psi_k'(t)}{2\pi}, \\
\qquad \psi_k'(t)&=\eta_{k,t}, \label{eq:syn-finst}
\end{aligned}
\end{equation}
i.e.\ a deterministic excursion $F(1\pm\varepsilon)$ plus $m\psi_k'/(2\pi)$ for
the $m$-th harmonic. Both terms are shared within the pair, so they cancel in
differences and the center-frequency separation is exactly
$\delta f_0(1+\varepsilon s(t))$.

\paragraph{Instantiation and splits.}
Each dataset has $T=10240$ steps, split chronologically $7{:}1{:}2$
($7168/1024/2048$), with input and horizon $96$; variables are standardized using
train-split statistics and metrics are reported in normalized space. We generate
five datasets at $\delta\in\{0,0.2,0.4,0.8\}$. Note that $\delta=0$ is a
no-discrepancy control, not a strictly stationary one: the random-walk phase
\eqref{eq:syn-walk} is drawn once and shared by all four settings, so it does not
drive any contrast.

\paragraph{Spectral verification.}
A model-agnostic check uses $2048$-step segments (resolution
$1/2048\approx4.9\times10^{-4}$ cycles/step), takes the dominant frequency in the
carrier band ($f>0.02$), and uses the $10/90$ percentiles over the series to
define a band midpoint and width. A variable passes if the midpoint is within
$\max(0.002,\,0.3\cdot2\varepsilon F_{k,j})$ cycles/step of its injected center
frequency; the $0.002$ floor (about four bins) accommodates peak wandering from
the sweep. The check is applied to all $80$ variable-level series
($20\times4$). Within-pair gaps are exact by construction ($\delta f_0$) and
compared against injected rather than estimated frequencies, since a $2048$-step
segment cannot resolve $\delta f_0$ for small $\delta$.



\section{Why Phase Hurts Optimization}
\label{app:phase_optimization}

\subsection{Local Sensitivity of Phase Representation}
Recall that the phase component is defined based on the ratio of the imaginary part to the real part:
\begin{equation}
\mathbf{Phase}_m = \arctan\!\left(\frac{\mathbf{f}_m^{im}}{\mathbf{f}_m^{re}}\right) \in \left[ -\frac{\pi}{2}, \frac{\pi}{2} \right].
\end{equation}
To analyze the local sensitivity, we consider a scalar element and let $x = f^{re}$, $y = f^{im}$, and $\mathrm{Amp} = \sqrt{x^2 + y^2}$. The partial derivatives of the phase with respect to the real and imaginary components are:
\begin{equation}
\frac{\partial \mathrm{Phase}}{\partial f^{re}} = \frac{1}{1 + (y/x)^2} \cdot \left( -\frac{y}{x^2} \right) = -\frac{f^{im}}{\mathrm{Amp}^2},
\end{equation}
\begin{equation}
\frac{\partial \mathrm{Phase}}{\partial f^{im}} = \frac{1}{1 + (y/x)^2} \cdot \left( \frac{1}{x} \right) = \frac{f^{re}}{\mathrm{Amp}^2}.
\end{equation}

\subsection{Gradient Instability in Low-Amplitude Regimes}
The above expressions reveal that the gradients of $\mathrm{Phase}$ are inversely proportional to the squared amplitude $\mathrm{Amp}^2$. Consequently, when $\mathrm{Amp}$ is small, the gradients of $\mathrm{Phase}$ with respect to $f^{real}$ and $f^{im}$ can become arbitrarily large.

This phenomenon frequently occurs during training, particularly at early stages when the magnitude of $\mathbf{f}_m$ is not yet well-structured. In such low-amplitude regimes, small perturbations in $f^{real}$ or $f^{im}$ induce disproportionately large changes in $\mathbf{Phase}_m$, leading to unstable gradient updates.

\subsection{Manifold Mismatch Between Phase and Neural Parameterization}
Beyond local gradient instability, phase variables exhibit an inherent geometric mismatch with standard neural network parameterizations. While $\mathbf{Phase}_m$ lies on a circular manifold, neural networks typically operate in Euclidean space and assume linear orderings of features.

As a result, values near the boundaries of the phase interval
\(
\left[-\frac{\pi}{2}, \frac{\pi}{2}\right]
\)
are numerically distant despite being geometrically adjacent on the circle. This discontinuity introduces artificial sharp transitions in the loss landscape, further complicating optimization.

\subsection{Weak Supervision of Phase Variables}
In typical sequence prediction tasks, the learning objective does not explicitly constrain the absolute phase $\mathbf{Phase}_m$. Instead, supervision is provided on aggregated or time-shift invariant quantities derived from the sequence.
As a consequence, phase variables introduce additional degrees of freedom that are weakly supervised or entirely unidentifiable from the training signal. Gradient-based optimization in such settings tends to exploit phase directions that reduce training loss without improving generalization, resulting in overfitting and degraded performance.

\subsection{Comparison with Amplitude-Based Optimization}
The amplitude component $\mathbf{Amp}_m$ exhibits well-behaved optimization properties. Its gradients with respect to $f^{real}$ and $f^{im}$ are given by
\begin{equation}
\frac{\partial \mathrm{Amp}}{\partial f^{real}} = \frac{f^{real}}{\mathrm{Amp}},
\quad
\frac{\partial \mathrm{Amp}}{\partial f^{im}} = \frac{f^{im}}{\mathrm{Amp}},
\end{equation}
which remain bounded except at the origin.

Moreover, $\mathbf{Amp}_m$ encodes a translation-invariant statistic that is directly aligned with the task objective, resulting in stable optimization dynamics and improved generalization.

\begin{figure}[h!]
 \centering
 \subfigure[MSE varying with the number of frequency-gated hidden states update blocks, the hidden frequency dimensions, and the learning rate.]{\includegraphics[width=1\linewidth]{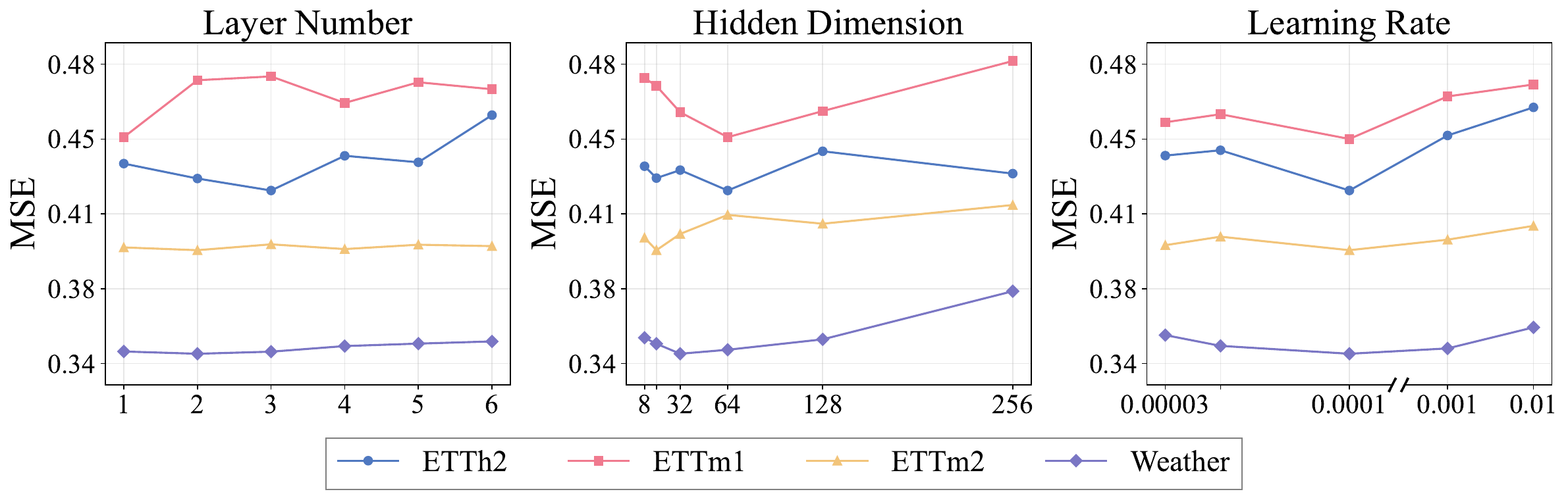}}
 \hfil
 \centering
 \subfigure[MAE varying with the number of frequency-gated hidden states update blocks, the hidden frequency dimensions, and the learning rate.]{\includegraphics[width=1\linewidth]{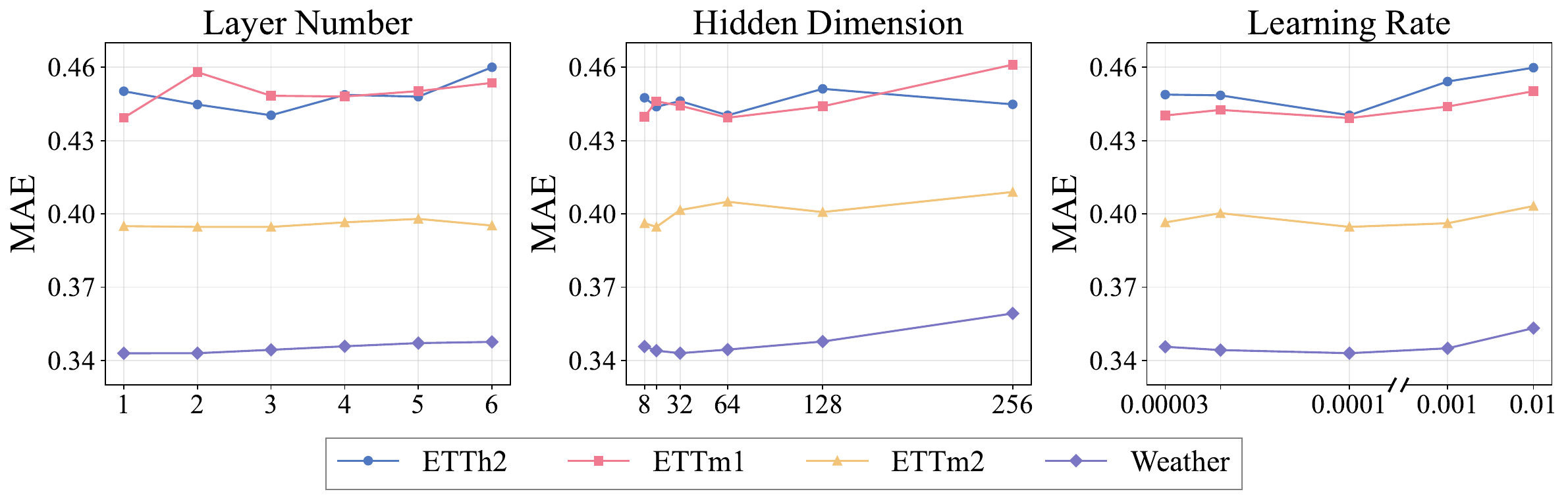}}
 \caption{Analysis of the hyperparameter sensitivity.}\label{fig:hyperparam_plots}
\end{figure}


\begin{figure}[h!]
    \centering
    \includegraphics[
        width=\linewidth
    ]{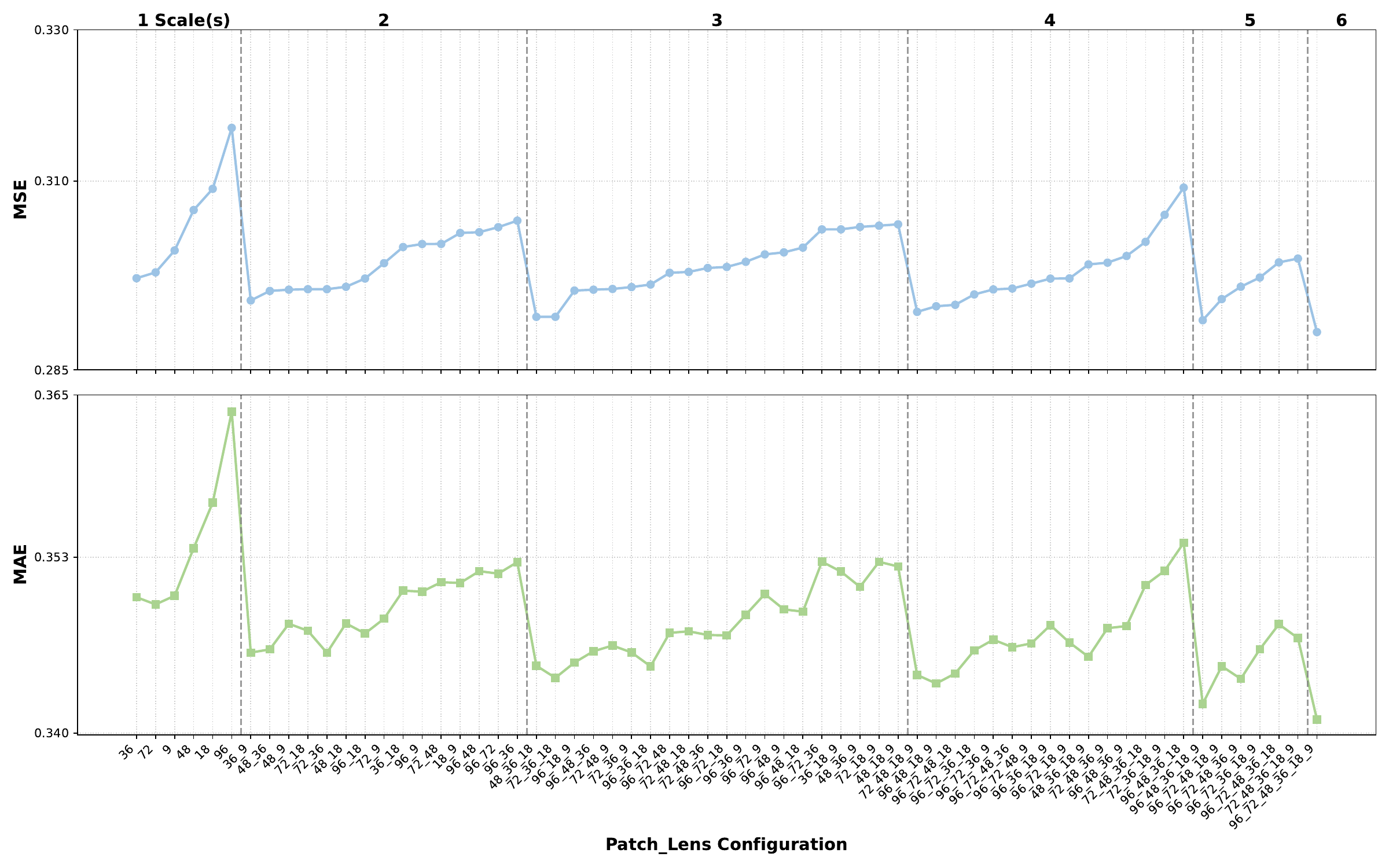}
    \caption{Sensitivity analysis of multi-scale patch configurations on the ETTh2 dataset.}
    \label{fig:multi_scale_patch_plots}
\end{figure}

\renewcommand{\arraystretch}{0.82}
\begin{table*}[h!]
\caption{The results on seven datasets with input length $T=96$ and prediction lengths $H \in \{96, 192, 336, 720\}$ (for ILI dataset, input length $T=36$ and prediction lengths $H \in \{24, 36, 48, 60\}$ ). \textbf{Bold results} are the best, and \underline{underlined results} are the second best.}
\centering
\footnotesize
\begin{tabular}{
>{\centering\arraybackslash}m{0.5cm}
@{\hspace{0.1pt}}>{\hfill}m{0.6cm}@{\hspace{2pt}}|
>{\centering\arraybackslash}m{0.5cm} >{\centering\arraybackslash}m{0.5cm} |
>{\centering\arraybackslash}m{0.5cm} >{\centering\arraybackslash}m{0.5cm} | 
>{\centering\arraybackslash}m{0.5cm} >{\centering\arraybackslash}m{0.5cm} | 
>{\centering\arraybackslash}m{0.5cm}>{\centering\arraybackslash}m{0.5cm} | 
>{\centering\arraybackslash}m{0.5cm} >{\centering\arraybackslash}m{0.5cm} | 
>{\centering\arraybackslash}m{0.5cm} >{\centering\arraybackslash}m{0.5cm} | 
>{\centering\arraybackslash}m{0.5cm} >{\centering\arraybackslash}m{0.5cm} | 
>{\centering\arraybackslash}m{0.5cm} >{\centering\arraybackslash}m{0.5cm} | 
>{\centering\arraybackslash}m{0.5cm} >{\centering\arraybackslash}m{0.5cm}  
}
\hline
\multicolumn{2}{c}{\textbf{Models}} 
& \multicolumn{2}{c|}{\makecell{\textbf{\M}\\\textbf{Ours}}} 
& \multicolumn{2}{c|}{\makecell{\textbf{Affirm}\\(2025)}} 
& \multicolumn{2}{c|}{\makecell{\textbf{S-Mamba}\\(2025)}} 
& \multicolumn{2}{@{}c@{}|}{\makecell{\textbf {STDMamba}\\(2025)}} 
& \multicolumn{2}{@{}c@{}|}{\makecell{\textbf {TimeMachine}\\(2024)}} 
& \multicolumn{2}{@{}c@{}|}{\makecell{\textbf{FreTS}\\(2023)}}
& \multicolumn{2}{@{}c@{}|}{\makecell{\textbf{PatchTST}\\(2023)}} 
& \multicolumn{2}{@{}c@{}|}{\makecell{\textbf{iTransformer}\\(2024)}}
& \multicolumn{2}{@{}c@{}}{\makecell{\textbf{ModernTCN}\\(2024)}} \\ 
\cmidrule(lr){1-2} \cmidrule(lr){3-4} \cmidrule(lr){5-6} \cmidrule(lr){7-8} \cmidrule(lr){9-10} \cmidrule(lr){11-12} \cmidrule(lr){13-14} \cmidrule(lr){15-16} \cmidrule(lr){17-18} \cmidrule(lr){19-20}
\multicolumn{2}{c}{\textbf{Metric}} & \textbf{MSE} & \textbf{MAE}  
& \textbf{MSE} & \textbf{MAE} & \textbf{MSE} & \textbf{MAE} & \textbf{MSE} & \textbf{MAE} 
& \textbf{MSE} & \textbf{MAE} & \textbf{MSE} & \textbf{MAE} & \textbf{MSE} & \textbf{MAE} & \textbf{MSE} & \textbf{MAE} & \textbf{MSE} & \textbf{MAE} \\ \hline
\multirow{5}{*}{\rotatebox{90}{\textbf{ETTh1}}} &  96  & \textbf{0.373} & 0.398 & 0.381 & \underline{0.395} & 0.386 & 0.405 & 0.384 & 0.408 & \underline{0.375} & \textbf{0.394} & 0.390 & 0.404  & 0.414 & 0.419 & 0.386 & 0.405 & 0.410 & 0.426 \\
& 192 & \textbf{0.428} & \underline{0.433} & \underline{0.430} & \textbf{0.426} & 0.443 & 0.437 & 0.434 & 0.435 & 0.435 & 0.460 & 0.445 & 0.439 & 0.460 & 0.445 & 0.441 & 0.436 & 0.466 & 0.458\\
& 336 & \underline{0.473} & 0.455 & 0.474 & \textbf{0.451} & 0.489 & 0.468 & \textbf{0.469} & \underline{0.454} & 0.502 & 0.467 & 0.501 & 0.470 & 0.501 & 0.466 & 0.487 & 0.458 & 0.510 & 0.481 \\
& 720 & \textbf{0.490} & \textbf{0.474} & \underline{0.494} & \underline{0.481} & 0.502 & 0.489 & 0.498 & 0.490 & 0.533 & 0.498 & 0.559 & 0.535 & 0.500 & 0.488 & 0.503 & 0.491 & 0.527 & 0.505 \\ 
& Avg & \textbf{0.441} & \underline{0.440} & \underline{0.445} & \textbf{0.438} & 0.455 & 0.450 & 0.446 & 0.447 & 0.461 & 0.447 & 0.475 & 0.462 & 0.469 & 0.454 & 0.454 & 0.447 & 0.478 & 0.468\\
\hline \midrule 
\multirow{5}{*}{\rotatebox{90}{\textbf{ETTh2}}} & 96  & \textbf{0.290} & \textbf{0.341} & 0.295 & \underline{0.345} & 0.296 & 0.348 & 0.300 & 0.394 & \underline{0.291} & \textbf{0.341} & 0.317 & 0.373 & 0.302 & 0.348 & 0.297 & {0.349} & 0.316 & 0.363\\
& 192 & \textbf{0.363} & \textbf{0.389} & \underline{0.368} & \underline{0.393} & 0.376 & 0.396 & 0.391 & 0.408 & 0.369 & 0.394 & 0.427 & 0.442 & {0.388} & {0.400} & 0.380 & 0.400 & 0.397 & 0.422 \\
& 336 & \underline{0.412} & \underline{0.425} & 0.416 & 0.428 & 0.424 & 0.431 & 0.423 & 0.431 & \textbf{0.407} & \textbf{0.422} & 0.526 & 0.505 & {0.426} & {0.433} & 0.428 & 0.432 & 0.443 & 0.447  \\
& 720 & \underline{0.420} & \textbf{0.440} & \textbf{0.417} & \textbf{0.440} & 0.426 & 0.444 & 0.432 & 0.445 & 0.426 & 0.443 & 0.684 & 0.591 & 0.431 & {0.446} & 0.427 & 0.445 & 0.464 & 0.469  \\ 
& Avg & \textbf{0.371} & \textbf{0.399} & 0.374 & 0.402 & 0.381 & 0.405 & 0.387 & 0.420 & \underline{0.373} & \underline{0.400} & 0.489 & 0.478 & {0.387} & {0.407} & 0.383 & 0.407 & 0.405 & 0.425 \\
\hline \midrule 
\multirow{5}{*}{\rotatebox{90}{\textbf{ETTm1}}} & 96  & \textbf{0.318} & \underline{0.361}  & \underline{0.324} & 0.363 & 0.333 & 0.368 & 0.365 & 0.392 & 0.329 & \textbf{0.358} & 0.335 & 0.371 & 0.329 & 0.367 & 0.334 & 0.368 & 0.333 & 0.372 \\
& 192 & \textbf{0.360} & \underline{0.381}  & 0.374 & 0.389 & 0.376 & 0.390 & 0.399 & 0.408 & \underline{0.363} & \textbf{0.380} & 0.377 & 0.394 & 0.367 & 0.385 & 0.377 & 0.391 & 0.367 & 0.389  \\
& 336 & \textbf{0.390} & \textbf{0.404}  & 0.407 & 0.409 & 0.408 & 0.413 & 0.434 & 0.428 & 0.398 & \underline{0.405} & 0.413 & 0.418 & 0.399 & {0.410} & 0.426 & 0.420 & \underline{0.397} & 0.410  \\
& 720 & \textbf{0.445} & \textbf{0.439}  & 0.468 & 0.446 & 0.475 & 0.488 & 0.487 & 0.458 & 0.463 & \underline{0.447} & 0.483 & 0.461 & 0.454 & \textbf{0.439} & 0.491 & 0.459 & \underline{0.460} & 0.448  \\ 
& Avg & \textbf{0.378} & \textbf{0.396} & 0.393 & 0.402 & 0.398 & 0.405 & 0.421 & 0.422 & \underline{0.388} & \underline{0.398} & 0.402 & 0.411 & 0.387 & {0.400} & 0.407 & 0.410 & 0.389 & 0.405 \\
\hline \midrule 
\multirow{5}{*}{\rotatebox{90}{\textbf{ETTm2}}} & 96  & \textbf{0.173} & \textbf{0.257} & 0.178 & 0.264 & 0.179 & 0.263 & 0.182 & 0.267 & 0.178 & {0.260} & 0.181 & 0.269 &\underline{0.175} & \underline{0.259} & 0.180 & 0.264 & {0.177} & 0.263 \\
& 192 & \textbf{0.237} & \textbf{0.299} & 0.247 & 0.306 & 0.250 & 0.309 & 0.250 & 0.308 & 0.245 & 0.304 & 0.249 & 0.322 & \underline{0.241} & \underline{0.302} & 0.250 & 0.309 & 0.248 & 0.310  \\
& 336 & \textbf{0.296} & \textbf{0.337} & 0.311 & 0.347 & 0.312 & 0.349 & 0.319 & 0.352 & 0.306 & 0.346 & 0.340 & 0.382 & \underline{0.305} & \underline{0.343} & 0.311 & 0.348 & 0.317 & 0.353  \\
& 720 & \textbf{0.393} & \textbf{0.394} & 0.415 & 0.410 & 0.411 & 0.406 & 0.420 & 0.409 & \underline{0.396} & \underline{0.395} & 0.449 & 0.455 & 0.402 & 0.400 & 0.412 & 0.407 & 0.423 & 0.415  \\ 
& Avg & \textbf{0.275} & \textbf{0.322} & 0.288 & 0.332 & 0.288 & 0.332 & 0.293 & 0.334 & \underline{0.281} & \underline{0.326} & 0.305 & 0.357 & \underline{0.281} & \underline{0.326} & 0.288 & 0.332 & 0.291 & 0.335 \\
\hline \midrule 
\multirow{5}{*}{\rotatebox{90}{\textbf{ILI}}} & 24  & 2.078 & \underline{0.854} & \underline{2.025} & 0.919  & 2.431 & 1.029  & 2.213 & 0.892  & \textbf{1.929} & \textbf{0.832}  & 2.797 & 1.098 & 2.159 & 0.893 & 2.136 & 0.893 & 2.108 & 0.882   \\
& 36 & \textbf{1.377} & \textbf{0.750} & 1.847 & 0.903  & 1.679 & 0.847  & \underline{1.664} & \underline{0.826}  & 2.030 & 0.862  & 2.882 & 1.124 & 2.349 & 0.924 & 2.187 & 0.894 & 2.329 & 0.939 \\
& 48 & \underline{1.870} & \underline{0.834} & 2.233 & 0.978  & 2.086 & 0.939  & 1.983 & 0.889  & \textbf{1.803} & \textbf{0.827}  & 2.841 & 1.133 & 2.137 & 0.913 & 2.032 & 0.870 & 2.207 & 0.898 \\
& 60 & \textbf{1.376} & \textbf{0.769}  & 1.986 & 0.922  & 2.046 & 0.916  & 1.843 & 0.866  & \underline{1.808} & \underline{0.847}  & 2.928 & 1.146 & 2.063 & 0.920 & 1.991 & 0.893 & 1.969 & 0.899 \\ 
& Avg & \textbf{1.675} & \textbf{0.802} & 2.023 & 0.930  & 2.060 & 0.933  & 1.926 & 0.868  & \underline{1.892} & \underline{0.842}  & 2.862 & 1.125 & 2.177 & 0.913 & 2.087 & 0.888 & 2.153 & 0.905 \\
\hline \midrule 
\multirow{5}{*}{\rotatebox{90}{\textbf{Weather}}} & 96  & \textbf{0.160} & \textbf{0.205} & 0.176 & 0.218 & \underline{0.165} & \underline{0.210} & 0.188 & 0.229 & 0.177 & 0.215 & 0.182 & 0.236 & 0.177 & 0.218 & 0.174 & 0.214 & 0.179 & 0.218 \\
& 192  & \textbf{0.208} & \textbf{0.249} & 0.224 & 0.258 & \underline{0.214} & \underline{0.252} & 0.234 & 0.266 & 0.224 & 0.256 & 0.219 & 0.271 & 0.225 & 0.259 & 0.221 & 0.254 & 0.225 & 0.259 \\
& 336  & \textbf{0.265} & \textbf{0.290} & 0.280 & 0.298 & 0.274 & {0.297} & 0.292 & 0.308 & 0.280 & 0.298 & \underline{0.270} & 0.313 & 0.278 & 0.297 & 0.278 & \underline{0.296} & 0.279 & {0.297} \\
& 720  & \textbf{0.344} & \textbf{0.342} & 0.357 & 0.348 & 0.350 & \underline{0.345} & 0.371 & 0.357 & 0.357 & 0.347 & \underline{0.348} & 0.380 & 0.354 & 0.348 & 0.358 & 0.347 & 0.356 & 0.346 \\ 
& Avg & \textbf{0.244} & \textbf{0.272} & 0.260 & 0.280 & \underline{0.251} & \underline{0.276} & 0.271 & 0.290 & 0.260 & 0.279 & 0.255 & 0.300 & 0.259 & 0.281 & 0.258 & 0.278 & 0.260 & 0.280 \\
\hline \midrule 
\multirow{5}{*}{\rotatebox{90}{\textbf{Exchange}}} & 96  & \textbf{0.080} & \textbf{0.197} & \underline{0.082} & \underline{0.199} & 0.086 & 0.207 & 0.093 & 0.214  & \underline{0.082} & 0.200 & 0.085 & 0.211 & {0.088} & {0.205} & 0.086 & 0.206 & 0.084 & 0.202 \\
& 192 & \textbf{0.167} & \textbf{0.291} & \underline{0.173} & \underline{0.295}  & 0.182 & 0.304 & 0.189 & 0.312  & 0.179 & 0.299 & 0.181 & 0.312 & 0.176 & 0.299 & 0.177 & 0.299 & 0.174 & \underline{0.295} \\
& 336 & \underline{0.315} &\underline{0.407} & 0.327 & 0.413 & 0.332 & 0.418 & 0.329 & 0.417  & 0.317 & 0.408 & 0.471 & 0.508 & \textbf{0.301} & \textbf{0.397} & 0.331 & 0.417 & 0.318 & \underline{0.407} \\
& 720 & 0.815 & 0.679 & 0.817 & 0.678 & 0.867 & 0.703 & 0.821 & 0.681  & \underline{0.812} & \underline{0.675} & 0.858 & 0.695 & {0.901} & 0.714 & 0.847 & 0.691 & \textbf{0.809} & \textbf{0.674} \\ 
& Avg & \textbf{0.344} & \textbf{0.394} & 0.350 & \underline{0.396} & 0.367 & 0.408 & 0.358 & 0.406 & 0.348 & \underline{0.396} & 0.399 & 0.432 & {0.367} & {0.404} & 0.360 & 0.403 & \underline{0.346} & \textbf{0.394} \\
\hline \midrule 
\multicolumn{2}{c|}{\textbf{1st Count}} & \textbf{28} & \textbf{22} & 1 & 4 & 0 & 0 & 1 & 0 & \underline{3} & \underline{7} & 0 & 1 & {1} & {2} & 0 & 0 & 1 & 2 \\ \hline
\end{tabular}

\label{tab:main_results_per}
\end{table*}

\section{Hyperparameter Sensitivity}\label{hyper_para}
To assess the sensitivity of our method to hyperparameter settings, we conducted experiments by varying the number
of frequency-gated hidden state update blocks $ F_n \in\left \{ 1,2,3,4,5,6 \right \} $, the hidden frequency dimensions $S \in\left \{ 8,16,32,64,128,256 \right \} $, and learning rates $lr \in\left \{ 3e-5,5e-5,1e-4,1e-3,1e-2 \right \} $ on the ETTh2, ETTm1, ETTm2 and Weather datasets. 
As illustrated in Fig.~\ref{fig:hyperparam_plots}, exerts the most pronounced influence on model performance and therefore requires careful tuning. 
Model performance generally improves with increasing hidden dimensions and remains relatively stable across different numbers of frequency-gated hidden states update blocks.

Furthermore, we conduct an exhaustive combinatorial sensitivity analysis on multiscale patch lengths $m_p$, evaluating 63 configurations (\textit{all combinations of 1 to 6 scales} from $\left \{96, 72, 48, 36, 18, 9 \right \} $). As illustrated in Fig.~\ref{fig:multi_scale_patch_plots}, single-scale baselines are highly sensitive to spectrum truncation, either missing macro-trends or over-smoothing volatility. Conversely, multiscale ensembles consistently yield robust, superior performance. By concatenating diverse patch embeddings, AdaMamba inherently captures a comprehensive spectrum, mathematically desensitizing the model to individual poor window choices.

\section{Full Forecasting Results.}
\label{full_pred}
Table~\ref{tab:main_results_per} reports the complete per-horizon results on the
seven public benchmarks, including MSE and MAE for each prediction length and
their averages; the best and second-best entries are highlighted in bold and
underlined, respectively. \M~obtains the best or tied-best average MSE on all seven datasets and the best average MAE on six.


\end{document}